\documentclass[lettersize,journal]{IEEEtran}
\usepackage{amsmath,amsfonts}
\usepackage{algorithmic}
\usepackage{algorithm}
\usepackage{array}
\usepackage[caption=false,font=normalsize,labelfont=sf,textfont=sf]{subfig}
\usepackage{textcomp}
\usepackage{stfloats}
\usepackage{url}
\usepackage{verbatim}
\usepackage{graphicx}
\usepackage{cite}
\usepackage[per-mode=symbol]{siunitx}
\usepackage{gensymb}
\usepackage{microtype}
\usepackage{xcolor}
\usepackage[font=small]{caption}
\usepackage{subcaption} 
\usepackage{tabularray}
\usepackage{graphics}
\usepackage{booktabs}
\usepackage{multirow}
\usepackage{threeparttable}
\usepackage{overpic}
\usepackage{makecell}
\usepackage{microtype}
\usepackage{hyperref}

\begin{document}

\title{ Learning thin deformable object manipulation \\ 
with a multi-sensory integrated soft hand}

\author
{Chao Zhao$^{\ast}$, Chunli Jiang$^{\ast}$, Lifan Luo$^{\ast}$, Shuai Yuan, Qifeng Chen, Hongyu Yu

\thanks{Manuscript received: November, 18, 2024; Revised April, 20, 2025; Accepted June, 17, 2025. This work was supported by the HKSAR (project: ITS/036/21FP) and the Foshan HKUST (project: FSUST21-HKUST03B, FSUST21- FYTRI04B). This article was recommended for publication by Editor Jeannette Bohg upon evaluation of the reviewers’ comments. (\textit{Corresponding authors: Qifeng Chen and Hongyu Yu.})
}

\thanks{$\ast$Authors with equal contribution. C. Zhao is with the School of Artificial Intelligence, Jilin University (email: {\tt\footnotesize chaozhao@jlu.edu.cn}). Part of this work was conducted during his time at the Hong Kong University of Science and Technology. C. Jiang, L. Luo, S. Yuan, Q. Chen, and H. Yu are with the Hong Kong University of Science and Technology, Clear Water Bay, Hong Kong (emails: {\tt\footnotesize \{lluoan, cjiangab, syuanaf\}@connect.ust.hk}, {\tt\footnotesize \{cqf, hongyuyu\}@ust.hk}).}

}

\markboth{Journal of \LaTeX\ Class Files,~Vol.~14, No.~8, August~2021}%
{Shell \MakeLowercase{\textit{et al.}}: A Sample Article Using IEEEtran.cls for IEEE Journals}

\maketitle

\begin{abstract}
Robotic manipulation has made significant advancements, with systems demonstrating high precision and repeatability. However, this remarkable precision often fails to translate into efficient manipulation of thin deformable objects. Current robotic systems lack imprecise dexterity, the ability to perform dexterous manipulation through robust and adaptive behaviors that do not rely on precise control. This paper explores the singulation and grasping of thin, deformable objects. Here, we propose a novel solution that incorporates passive compliance, touch, and proprioception into thin, deformable object manipulation. Our system employs a soft, underactuated hand that provides passive compliance, facilitating adaptive and gentle interactions to dexterously manipulate deformable objects without requiring precise control. The tactile and force/torque sensors equipped on the hand, along with a depth camera, gather sensory data required for manipulation via the proposed slip module. The manipulation policies are learned directly from raw sensory data via model-free reinforcement learning, bypassing explicit environmental and object modeling. We implement a hierarchical double-loop learning process to enhance learning efficiency by decoupling the action space. Our method was deployed on real-world robots and trained in a self-supervised manner. The resulting policy was tested on a variety of challenging tasks that were beyond the capabilities of prior studies, ranging from displaying suit fabric like a salesperson to turning pages of sheet music for violinists.

\end{abstract}


\section{Introduction}

Thin, deformable objects are in many daily activities. For robots to be widely used in human living environments, they must be able to manipulate these objects. Despite notable progress in robotic manipulation \cite{fp0,fp6, zhu2022challenges} and the deployment of numerous commercial robotic platforms, such as Covariant’s AI-driven picking solutions \cite{fp7} and Boston Dynamics' Stretch warehouse robot \cite{fp9}, even the most advanced robots fall short of human performance when manipulating thin, deformable objects. We suggest that advancements in thin deformable object manipulation hinge on achieving “imprecise dexterity” — the ability to perform dexterous manipulation through robust and adaptive behaviors despite imprecision in control, movement, and perception.

\begin{figure}[!t]
    \centering
    \includegraphics[width=1\linewidth]{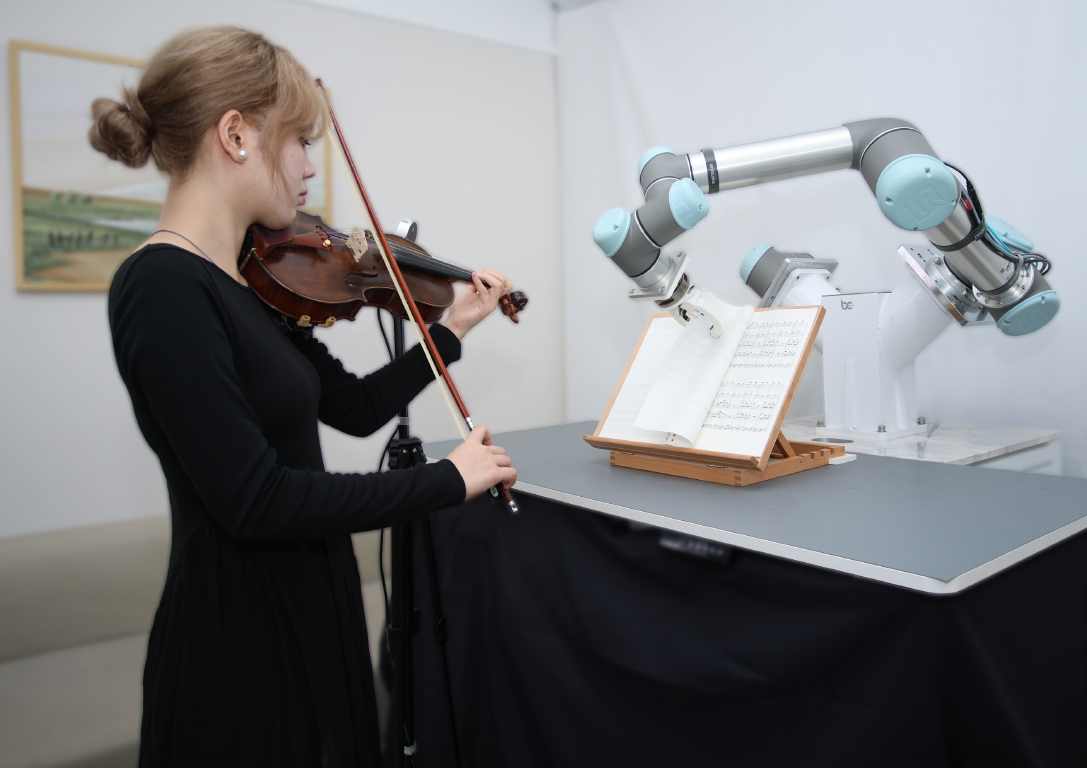}
    \caption{\textbf{Our robot assists a violinist with page-turning during performances.}}
    \label{fig:1}
\vspace{-18pt}
\end{figure}

Singulation and grasping thin, deformable objects are common daily tasks, such as turning book pages (Fig. \ref{fig:1}). These tasks present intricate dynamics due to the thin structures, stacking tendencies, and high-dimensional configuration spaces of the objects, posing significant challenges in manipulation control and perception. Conventional robots, which rely on precise control, struggle to meet multiple objectives: adapting to surface deformation and slipping, maintaining stable contact, avoiding object damage, and ensuring successful singulation and grasping \cite{l6}. Visual sensors often fail to capture essential physical properties like friction and stiffness, and can be occluded by fingers or objects, necessitating heavy reliance on touch and proprioception for feedback on object interaction. While humans effortlessly perform such tasks with hands that integrate passive compliance, touch, and proprioception  \cite{m4}, it remains an open challenge in robotics. 

Previous work has not demonstrated robust systems for the singulation and grasping of thin, deformable objects, often being limited to a single object type \cite{l3, flipbot, revision_R2_1, revision_R2_2} or assuming an isolated object within the environment \cite{revision_R1_1}. In downstream applications such as robot-assisted dressing \cite{k1, k2}, garment folding \cite{k4}, and bagging \cite{k3}, researchers tend to overlook this critical step, often making assumptions about grasping or relying on heuristic strategies.

Here, we present a robust approach for singulating and grasping a variety of thin deformable objects in real-world scenarios that are beyond the capability of prior published work in robotics (Movie 1\footnote{All the videos are available at https://robotll.github.io/LTDOM/}). 

One of the key components is our soft, underactuated two-finger hand. Different thin deformable objects exhibit various behaviors under external forces; for instance, fabric dents easily while pages resist pressure but yield to shear forces. Our gripper's passive compliance allows its fingers to follow the object's deformation, maintaining continuous contact and limiting potentially damaging forces exerted on both objects and the robot \cite{l6}. Such adaptability enables gentle, force-limited interactions, resulting in successful grasps without requiring perfect robot precision \cite{revision5, revision_R2_3}. Compared with the gripper in Flipbot \cite{flipbot}, we maintained its general design but augmented its sensory capabilities. Our hand integrates tactile sensors at each fingertip alongside the original F/T sensors, offering passive compliance, touch, and muscle proprioception similar to a human hand. The addition of tactile sensing enables the robot to perceive local contact features closely linked to material properties such as texture and hardness. In contrast, Flipbot’s reliance solely on global force information provided by F/T sensors limits its performance and generalization potential, restricting it to the paper-like object.

Like Flipbot \cite{flipbot}, we learned a direct mapping from perception to robot control using model-free RL in the real world to drive our soft hand. Model-based methods, while powerful in structured domains, struggle in deformable object manipulation due to the inherent uncertainty and complexity of thin deformable objects. Their high-dimensional configuration spaces, coupled with partially observable dynamics and large variations in material properties (e.g., bending stiffness, friction), pose significant challenges for analytical modeling. In contrast, model-free RL learn manipulation policies end-to-end from trial-and-error exploration, without requiring explicit modeling of object dynamics. However, model-free RL alone was insufficient for learning the robust policy as presented in our video. Therefore, we incorporated and validated several additional components to achieve the demonstrated capabilities. 

The first is a slip module. We found that the reward sparsity in these tasks makes robot impossible to discover effective policies from scratch, and acquiring task-relevant perception remains a problem. To address these, we employ an active exploratory motion called slip, where robot fingers slide over the object's surface. In Flipbot \cite{flipbot}, the slip motion is hardcoded and fixed for specific scenarios, making it impossible to generalize to different object sizes and positions. Now, slip motion is guided by a neural network that receives inputs from a wrist-mounted camera and is trained on data reflecting human intuition (e.g., turning book pages from the upper right). Such position-specific slips not only provide the robot with a proper initial pose, reducing the search space for policy learning, but also generate rich, dynamic interactions that expose latent physical properties like flexural rigidity and friction.

Another component is a hierarchical, dual-loop learning structure, designed to improve learning efficiency—an essential factor for real-world reinforcement learning. Flipbot \cite{flipbot} introduced a reset mechanism and an automatic reward signal based on the page number, significantly reducing human intervention and enabling policy learning within four hours. Building on this framework, we extend the training setup to include fabric materials. This increased variety necessitates greater exploration and longer training durations. To address this, the dual-loop structure accelerates learning through the structured and efficient exploration and exploitation of the action space. The necessary action range after slip motion varies due to differences in material surface properties and movement errors caused by depth camera noise. The outer loop selects between coarse and fine action spaces based on a required range of movement, while the inner loop outputs concrete robot actions. By selecting the appropriate action space based on current perception and learned experience, unnecessary action exploration is reduced.

The presented study achieves a significant leap over the existing work in terms of graspable object variety and performance, marking a meaningful step toward human-level sensorimotor intelligence. Notably, our training only uses printer paper and winter fabric on flat surfaces. Yet when deployed in environments, the robot successfully handles various deformable objects with different physical properties (coated paper, hotel towel,  summer suit fabric), hybrid materials (fabrics separated by the introduction page or spring pancakes separated by baking paper), and gravitational effects. The singulating and grasping of thin, deformable objects have long been difficult problems in robotics. Our results suggest that the extraordinary complexity of manipulating thin, deformable objects can be tamed with the integration of embodied multi-sensory integration and passive compliance.

The main contributions of this article are as follows:
\begin{enumerate}
    \item We propose a novel approach for singulation and grasping thin, deformable objects by controlling a soft, underactuated hand that integrates passive compliance and multi-modal perception. To the best of our knowledge, this is the first work to successfully demonstrate the singulation and grasping across such a wide range of thin deformable objects under challenging real-world conditions.
    \item We introduce a slip module that actively collects rich multisensory information and provides a favorable initial pose prior to manipulation. This enhances task awareness and learning performance.
    \item We present a dual-loop learning structure that adaptively selects action space granularity based on sensory input. The mechanism improves sample efficiency and accelerates policy convergence.
    \item We validate our approach through extensive experiments, including ablation studies, quantitative evaluations of passive compliance, and analysis of the underlying mechanisms enabling successful manipulation.
\end{enumerate}

This journal paper evolves upon the conference paper presented in \cite{flipbot}. Section \ref{sec: method} introduces the problem formulation (Section \ref{sec:method-pro}), a slip module (Section \ref{sec:method-slip}) for obtaining sensory information, and a double-loop learning structure. We conduct extensive experiments (Section \ref{sec:results}) demonstrating the effectiveness of our method, while also evaluating the role of passive compliance (Section \ref{sec:results-nx} and Section \ref{sec:results-compliance}) and multisensory integration (Section \ref{sec:results-Visualization}). A comprehensive component-wise analysis is provided (Section \ref{sec:results-Ablation}), along with a discussion on limitations and future opportunities (Section \ref{sec:results-Discussion}).

\section{Related Work}

Deformable object manipulation encompasses a range of strategies. One line of research involves model-based methods, which often revolve around accurately modeling the object's dynamics and subsequently applying model predictive control or trajectory optimization  \cite{a1,a4,a5}. These methods heavily depend on the precise information about the object and any uncertainties encoded in the dynamics model being accurate  \cite{a3, c1, l4}. In the context of thin, deformable objects, model-based approaches face significant challenges due to multiple sources of uncertainty. Accurately estimating physical properties such as bending stiffness and surface friction is inherently difficult, as these parameters vary even among seemingly identical objects due to differences in texture, wear, and manufacturing variability. Moreover, such objects exhibit high-dimensional and nonlinear configuration spaces, where small action perturbations result in drastically different deformations, rendering reliable forward prediction intractable. These uncertainties are further exacerbated by sensor occlusion, depth perception errors, and complex finger-object interactions, which are difficult to incorporate into analytical models.

Another line of research that has gained substantial attention for deformable object manipulation is model-free approaches. Notably, Navarro-Alarcon et al. \cite{navarro2013model} proposed a model-free visual servoing framework to control the deformation of elastic objects using closed-loop feedback, bypassing the need for explicit object modeling. Concurrently, learning-based methods have prominence in deformable object manipulation \cite{d1, d2}, which is largely attributed to the capability to infer action from sensory input, bypassing the complexities of direct physical modeling \cite{e1, e2, k5, f2}. 

Within this domain, model-free reinforcement learning (RL) methods \cite{ibarz2021train, ccalicsir2019model} have shown great potential for manipulating thin, deformable objects. Model-free RL features by learning to control robots directly from raw sensory data through continuous trial and error, introducing a paradigm shift in many robotic areas, including whole-body control for door opening \cite{h1}, planetary exploration using legged robots \cite{h2}, and autonomous flight of micro aerial swarms \cite{h3}. The manipulation of deformable objects using RL has made significant advances in environments with abundant simulation data \cite{i1, i3}. Moreover, sim-to-real transfer techniques have further extended these capabilities to real-world applications \cite{j1}. However, applying model-free RL to learn thin, deformable object manipulation remains challenging. These objects' sensitivity to external forces requires the common rigid robot hands to precisely control the contact between fingers and the object, as well as manage the amount and timing of the applied force \cite{l6}. This creates a high-dimensional control space that is difficult to learn with RL and poses risks of damaging both the robot and the object during real-world exploration. The challenge is exacerbated by the inability of simulators, commonly used to accelerate RL training, to replicate the dynamics of thin, deformable objects accurately. Blanco-Mulero et al. \cite{blanco2024benchmarking} systematically analyzed this gap, revealing that state-of-the-art simulators struggle to accurately replicate the nuanced physical behaviors of cloth. Successful learning to manipulate these objects may require simplifying manipulation while ensuring adaptability to varying object properties without relying on high-precision control. 

Soft robotic hands \cite{soft1, soft2, soft3} have recently gained attention for simplifying manipulation while adapting to varying object properties without the need for high-precision control \cite{abondance2020dexterous, l6}. Studies show that soft hands offer passive stability and make manipulations easier to achieve \cite{revision5, revision_R2_3}. Meanwhile, the inherent compliance of soft robotic hands handles uncertainties during object-fingertip engagement gracefully and limits the potentially damaging forces exerted on thin, deformable objects \cite{l6}. 

Complementing the control problem is acquiring the perception needed for singulating and grasping thin, deformable objects. Visual sensors, which most work relies on \cite{depth2, depth3, h3}, provide position and shape information but cannot detect inter-layer interactions or physical properties. Force/torque sensors, akin to human muscle proprioception \cite{revision3}, offer a comprehensive view of system forces and torques yet are limited to single-point contact data \cite{revision2, flipbot}. Recent studies have employed tactile sensors \cite{Gelsight, DIGIT} to measure local physical properties of deformable objects, such as texture, material, and thickness \cite{l1, l2}. However, tactile information alone is limited \cite{m4}, lacking a holistic view of the object’s entire structure and behavior under manipulation. Therefore, a combination of visual, force/torque, and tactile sensing is required to achieve a comprehensive perception for manipulating thin, deformable objects.

\begin{figure*}[!t]
    \centering
    \includegraphics[width=1\linewidth]{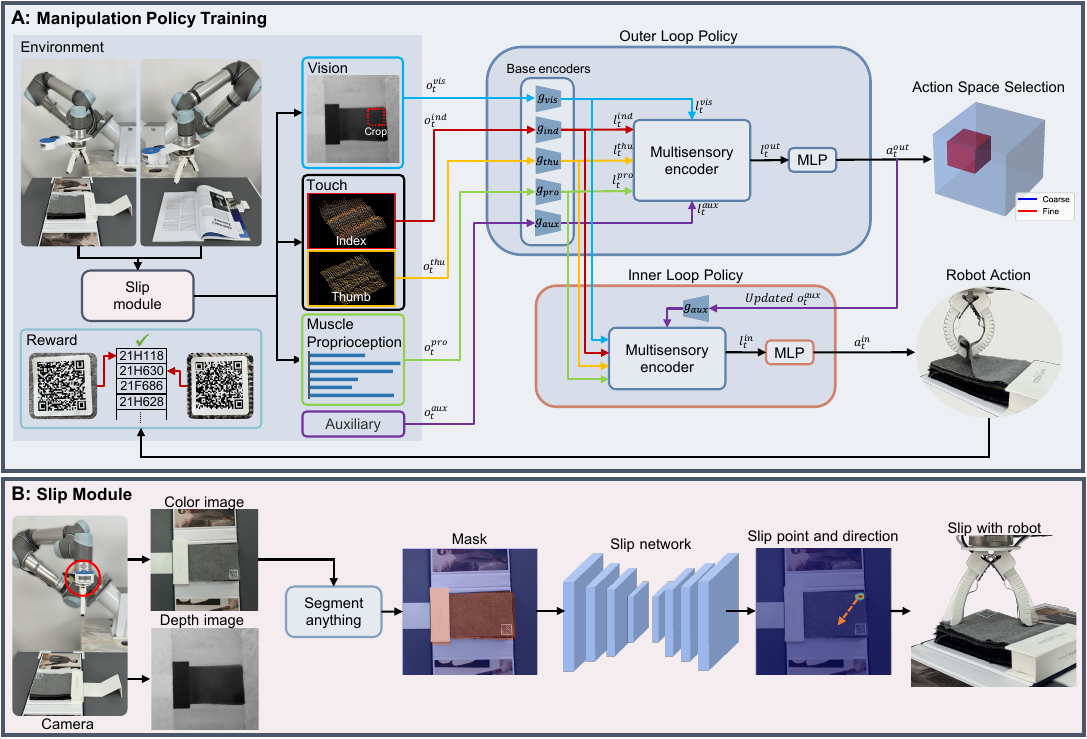}
    \caption{\textbf{Overview of the presented approach.} (\textbf{A}) We train manipulation policies in the real world using RL. During training, the robot proactively interacts with the object to obtain multisensory observations via a slip module.  The outer loop selects an action space based on observations, and the inner loop explores within it to output concrete actions for the robot to execute. After training, we deploy the learned policy zero-shot on robots for various tasks. (\textbf{B}) Slip module. The object mask, derived from the segment anything model, is fed into a fully convolutional network to infer pixel-wise affordance maps. The chosen pixel is decoded into slip motion, with its 3D coordinates derived from the depth image.}
    \label{fig:2}
\vspace{-18pt}
\end{figure*}

\section{Methods}\label{sec: method}

We learn a neural network policy on a real-world robotic platform to enable the robot to effectively grasp thin, deformable objects using a soft hand. Our method consists of the following stages, as shown in Fig. \ref{fig:2}A.

First, the robot obtains multisensory observations through the slip module (Fig. \ref{fig:2}B). The slip module is pre-trained using data reflecting human preferences. We use an RGB-D camera to obtain visual observations, allowing for the inference of the slip point and its direction by the slip module. Subsequently, the robot touches and slips with objects and monitors through tactile sensors and force-torque sensors. Thus, the obtained multisensory observations provide insights into unobservable physical information of the environment state. 

Then, the outer loop selects the action space from two granularity levels, coarse and fine, based on the multisensory observations. The coarse action space offers broader movements with sparse discretization, while the fine action space covers a smaller range but benefits from dense discretization. The outer loop utilizes the multisensory encoder to integrate the observations from different senses into a latent vector representation, which then guides the action space selection.

Finally, the inner loop explores and learns within the selected action space. Employing the same multisensory encoder, the inner loop maps the observations to the robot's concrete action. During training, the rewards obtained by the inner loop are shared with the outer loop, achieving mutual performance improvement. 

\subsection{Hardware setup}\label{sec:method-har}

\begin{figure*}
    \centering
    \includegraphics[width=1\linewidth]{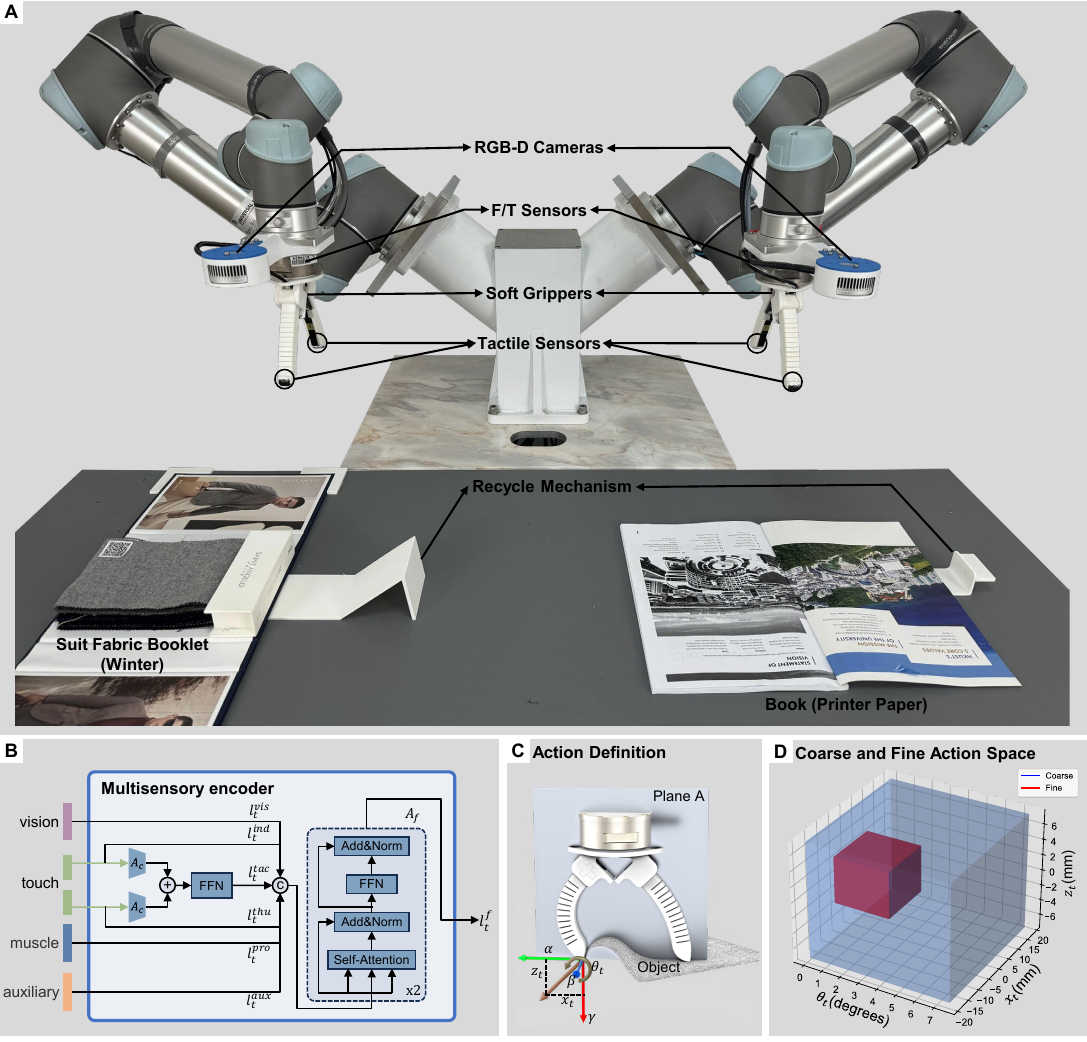}
    \caption{\textbf{Details of components for robust policy learning.} (\textbf{A}) Our hardware setup for policy learning. Two robots explore and learn policies simultaneously in parallel environments, updating their policies based on the rewards received. (\textbf{B}) Multisensory encoder architecture. It incorporates a cross-attention mechanism $g_c$ and a transformer encoder $g_\kappa$ that merges features from different senses into a latent vector. FFN denotes the feedforward network. (\textbf{C} and \textbf{D}) Visualization of our action coordinate system, as well as the coarse and fine action spaces.}
    \label{fig:3}
\vspace{-18pt}
\end{figure*}

Fig. \ref{fig:3}A provides an overview of the hardware setup used for policy learning. The setup includes two Universal Robot 5 robot arms, each equipped with an Intel RealSense L515  camera, an ATI gamma Force Torque sensor at the wrist joint, and a two-finger soft robotic hand. 

The soft robotic hand is based on the design of the fast-PneuNet \cite{revision_R1_4}, a simple pneumatic actuator widely used in the research community. The whole hand is made of thermoplastic polyurethane (TPU) using a consumer-grade 3D printer. The index and thumb fingertips of each hand are equipped with a vision-based tactile sensor \cite{OurTouch}.  Furthermore, the F/T sensor and the soft robotic hand are fixed together, moving and rotating in unison. Section \ref{sec:s4} provides further details about tactile sensors, and we discuss our gripper design in Section \ref{sec:s9}. 

We maneuver the robot using position control within the local/tool coordinate system $H_{\alpha\beta\gamma}$, which is established at the index fingertip, as shown in Fig. \ref{fig:3}C. The $\alpha$ axis is oriented along the line that connects the two fingertips and lies within the longitudinal plane $\Lambda$ formed by two fingers; the $\gamma$ axis is orthogonal to the $\alpha$ axis and remains within plane $\Lambda$; the $\beta$ axis is perpendicular to the plane formed by $\alpha$ and $\gamma$. Each command issued to the robot is interpreted as movements relative to its current pose. Compared with directly controlling the joint angles of the robot arm, this allows us to simplify the dimensions of the action and reduce the difficulty of policy learning. 

\subsection{Problem formulation}\label{sec:method-pro}
We formulate the problem of singulation and grasping thin, deformable objects as a Markov Decision Process (MDP), with a state space $S$, action space $A$, a scalar reward function $R$, and a transition probability $P(s_{t+1} |s_t, a_t)$. Given our model-free RL setup, the transition dynamics $P$ are not explicitly modeled but instead learned implicitly through trial-and-error interactions with the environment. At time step $t$, an agent uses a policy $\pi$ to select an action $a_t$ and receives rewards $r_t = R(s_t, a_t)$. 

Specifically, we adopt a hierarchical policy learning framework consisting of two interrelated loops: an outer loop policy $\pi^{out}$ and an inner loop policy $\pi^{in}$. Given the current state, the agent selects an action granularity $a^{out}$ using $\pi^{out}$. Then, given $a^{out}$, the agent predicts a concrete action $a^{in}$ for the robot to execute using $\pi^{in}$. The objective is to learn an optimal joint policy $(\pi^{out*}, \pi^{in*})$ that maximizes the expected cumulative reward:
\begin{multline}
(\pi^{out*}, \pi^{in*}) 
= \mathop{\arg \max}_{\pi^{out}, \pi^{in}} \; \\
E\bigg[ \sum_{t=0}^{\infty} \lambda^t \,
R\big( s_t, \pi^{in}(s_t \mid \pi^{out}(s_t)) \big) \bigg],
\end{multline}
where ${E}[\cdot]$ represents the expectation under the joint policy $(\pi^{out*}, \pi^{in*})$, and $\lambda$ is the discount factor. The reward function $R$ is binary, assigning a value of 1 for task success and 0 otherwise. We utilize Soft Actor-Critic (SAC) \cite{SAC} for learning the policies. The definitions of state, action, and reward are as follows.

\subsubsection{State}

The state for both inner and outer loop policy is defined as $s_t = (o_t^{vis}, o_t^{ind}, o_t^{thu}, o_t^{pro}, o_t^{aux})$, where $o_t^{vis}$ refers to the visual observation from the camera, $(o_t^{ind}, o_t^{thu})$ refers to the touch observation from the index and thumb fingertips, respectively; $o_t^{pro}$ refers to the muscle proprioception from the F/T sensor, and $o_t^{aux}$ refers to the auxiliary state. The auxiliary state includes the action space selection and a flag representing whether the current stage belongs to the inner or outer loop. In the outer loop, the flag is assigned a placeholder value of zero to maintain uniformity in the observation format, thereby facilitating the reuse of the multisensory encoder by the inner loop.

\subsubsection{Action}

After the observation is collected, the outer loop selects the action space $a^{out}$ from two different granularities, coarse and fine (Fig. \ref{fig:3}D). Within the selected action space, the inner loop predicts the action $a^{in}$ that the robot needs to perform. Specifically, the output of the inner loop is rescaled and mapped to different action ranges based on action space selection, and each coordinate of the action $a^{in}$ is discretized accordingly. 

Formally, the action $a_t^{in}$ is defined as $(x_t,z_t,\theta_t, \Omega)$, as shown in Fig. \ref{fig:3}C. Here, $(x_t,z_t,\theta_t)$ represents the gripper displacement, representing the relative discrepancy between the current and the desired gripper pose after the slip motion. Specifically, $x_t$ and $z_t$ are the displacement along axis $\alpha$ and $\gamma$, respectively; $\theta_t$ denotes the orientation of the gripper about the axis $\beta$. The parameter $x_t$ adjusts the distance of the index finger from the object's edge. Positioning closer to the edge reduces the required manipulation energy, but being too close or beyond the edge can cause multiple layers to flip up. The parameter $z_t$ regulates the pressure applied to the object's surface to ensure adequate friction during manipulation. The parameter $\theta_t$ enables the robot to distribute varying amounts of force between the two fingers, and the appropriate distribution will facilitate separation and grasping. Additionally, $\Omega$ is a Boolean vector that represents either the closing or opening of the pneumatic finger, which is set to true at the end of the episode to close the gripper. More details of the observation and action are in Section \ref{sec:s5}.

\subsubsection{Reward}

At the end of an episode, we provide a positive reward 1 for successfully grasping a single layer of an object. In contrast, results where zero, two or more layers are grasped simultaneously receive a reward of 0. The reward $r_t$ in both the book and fabric environments is determined automatically with information naturally possessed by the manipulated object. Specifically, rewards in the book environment are determined by referencing the page number, and in the fabric environment by matching QR code data to a pre-set sequence in the manufacturer's database. The episode length is fixed to one in order to learn a successful policy within a minimal number of steps. 

\subsection{Observations via slip module}\label{sec:method-slip}

The slip module serves two primary functions: obtaining task-related sensory information and providing an initial robot pose to minimize unnecessary exploration. To obtain rich multisensory observations, we employ a slip motion to proactively interact with the object (Fig. \ref{fig:2}B). This process begins with the robot capturing aligned RGB and depth images from above the object. The RGB image is processed through the Segment Anything Model (SAM) \cite{SAM} to generate object masks. These binary masks are rotated at various angles and subsequently input into the slip network. The slip network outputs a set of affordance maps, one for each input rotation angle. Each map contains values corresponding to the slip motion primitive, which are parameterized by the pixel's location and its rotation angle within the map. The pixel with the highest affordance value is selected; its spatial position identifies the slip point, and the rotation angle of the corresponding input mask determines the slip direction. 

Once the slip point and direction are determined, the depth image is utilized to ascertain the 3D coordinates of this selected pixel in the camera's frame. This information, combined with a pre-established hand-eye calibration relating the camera and robot fingertip coordinate systems, allows the robot to move to the slip point.

Next, the robot executes the slip motion. First, the robot's hand rotates such that the $\alpha$ axis (the line connecting the two fingertips) aligns with the slip direction. The plane composed of the $\alpha$ and $\beta$ axes is parallel to the workspace. We define the workspace as the planar supporting surface upon which objects are placed for manipulation experiments, physically corresponding to the table or platform in the robot's operational environment. During training, the workspace inclination remains consistently at 0 degrees. 

The index finger then moves to the predicted slip point, stopping at a standoff distance of 3 mm above the object's surface, with this height being estimated using the depth camera feed. Contact is then initiated by applying positive air pressure, causing the soft finger to inflate and bend downwards to slip onto the object's surface.  After 1 second, we record sensory information. The air pressure is then released, and the fingers relax. This 3 mm standoff distance was determined empirically; it provides a balance, being large enough to prevent unintended collisions due to minor sensing inaccuracies or positioning errors, yet small enough to ensure reliable surface contact and the acquisition of high-quality sensory data upon slipping. 

After a slip, we record 3D arrays of deformation flows from the tactile sensors as touch observation ($o_t^{ind}, o_t^{thu}$), along with readings from the F/T sensor as the muscle proprioception $o_t^{pro}$, including forces $(f{\alpha}, f{\beta}, f{\gamma})$ and torques $(m{\alpha}, m{\beta}, m{\gamma})$.   The three main axes of the F/T sensor are aligned with the three corresponding axes of the $H_{\alpha\beta\gamma}$. This alignment makes the force and torque data measured by the sensor more intuitive. Moreover, we crop the area surrounding the slip point on the depth image to serve as visual observation $o_t^{vis}$.  

Through the robot's purposive movements prior to manipulation, observations are made that maximize the information uptake on the relevant properties of the target object. These observations provide insights into the elaborate interplay between static and dynamic sensory cues, such as texture, softness, and friction. The slip network is pre-trained using synthetic data before the robot learns in the real world. Humans annotated slip points on a few images featuring a book, a shirt, and a spring pancake. We employed data augmentation techniques, including rotation, translation, and scaling, on the object masks of these images to simulate different object sizes and positions. More details of the slip network architecture are in Section \ref{sec:s6}.

\subsection{Policy architecture}

Both inner and outer loop policies are implemented by neural networks. We model the outer loop policy $\pi^{out}$ with the following components: basic encoders, a multisensory encoder, and a multilayer perceptron (MLP), as illustrated in Fig. \ref{fig:2}A. 

The basic encoders consist of the vision encoder $g_{vis}$, tactile encoders $(g_{ind}, g_{thu})$ for each finger, the encoder $g_{pro}$ for F/T measurement, and the auxiliary state encoder $g_{aux}$. Given the observation $o_t$, these encoders compress each sensory element into compact, single-dimensional feature vectors $(l_t^{vis}, l_t^{ind}, l_t^{thu}, l_t^{pro}, l_t^{aux})$.  This process enables the transformation of sensory data into a format that is compatible with the input of the multisensory encoder. 

The multisensory encoder $g_{mul}$ receives these feature vectors and fuses them into a smaller latent representation $l_t^{out}$:
\begin{align}
l_t^{out} =  g_{mul}(l_t^{vis}, l_t^{ind}, l_t^{thu}, l_t^{pro}, l_t^{aux}).
\end{align}
The MLP then uses the $l_t^{out}$ to output the selection of the action space. For the inner loop policy, we reuse the basic encoders and the multisensory encoder from the outer loop but with a different final MLP block to predict the action.

\subsection{Multisensory encoder}\label{sec:method-enc}

To integrate vision, touch, and F/T sensor data, we introduce the multisensory encoder, as shown in Fig. \ref{fig:3}B. The multisensory encoder $g_{mul}$ consists of a cross-attention block $g_{c}$ \cite{CrossAttention} and a transformer encoder $g_{\kappa}$ \cite{Attention}.

The tactile information of the index finger and thumb is complementary as it perceives invisible details about objects from different angles and positions. The cross-attention mechanism is adept at capturing features between such information that is related to each other and has dependencies \cite{CrossAttention}. Within the cross-attention block $g_{c}$, the touch information $l^{ind}_t$ and $l^{thu}_t$, is fused into a new latent vector $l^{c}_t$: 
\begin{align}
l^{c}_t = g_{c}(l^{ind}_t,l^{thu}_t).
\end{align}
Subsequently, $l^{c}_t$, along with the other latent vectors of different sensory information, serve as a sequence of inputs to the transformer encoder $g_{\kappa}$. It performs mutual fusion encoding among the different sensory data modalities to output a vector $l_t^{\kappa}$ that forms a compact representation for mapping actions:
\begin{align}
l_t^{\kappa} = g_{\kappa}(l_t^{\text{vis}}, l^{ind}_t,l^{thu}_t, l_t^{\text{c}}, l_t^{\text{pro}}, l_t^{\text{aux}}).
\end{align}
The multisensory encoder undergoes an implicit learning process during RL training, avoiding resorting to heuristic-based explicit analysis, which often falters across diverse environments. Network layer details about the multisensory encoder are in Section \ref{sec:s7}.

\subsection{Training details}

The training process of our method consists of two distinct phases: 1) training the slip module using synthetic data, and 2) learning the manipulation policy through reinforcement learning in the real world. 

\subsubsection{Slip module training} The slip module is independently trained through supervised learning prior to the RL policy training phase. We collected images of three representative objects: a printed book, a shirt, and a stack of spring pancakes. For each image, appropriate slip points and corresponding directions were manually annotated. To ensure robust generalization to various object positions and orientations, we performed extensive data augmentation, including random rotations, translations, and scaling transformations, generating a diverse synthetic dataset. The slip network is trained on this augmented dataset using the Adam optimizer with cross-entropy loss. Once trained, the slip network parameters are frozen and subsequently utilized for predicting slip points during the RL policy learning phase.

\subsubsection{Manipulation policy training via RL} In this phase, robots learn the manipulation policy in the real world (Fig. \ref{fig:3}A). Operating in parallel environments, robotic arms execute these tasks while continuously updating the policies according to rewards. Specifically, we train our model through trial and error with the following procedure: 

At the start of each training episode, two robotic arms use their wrist cameras to capture RGB-D images of environments, which are subsequently processed by the pretrained (and now frozen) slip module to predict the position and direction of the slip motion.  Following this prediction, the robot starts to perform a slip motion from its initial posture in each environment to obtain multisensory observations. Then, the robot downloads the latest parameters for both the inner and outer loop policies from the optimizer. The outer loop policy $\pi^{out}$ selects an action space based on multisensory observations. The inner loop policy $\pi^{in}$ explores within the range of the action space to output a concrete action. Rewards are obtained after robot execution. Finally, training episodes generated from both environments are added to the replay buffer, from which the optimizer samples to update the policy. Both the inner and outer loop policies are periodically updated using the Adam optimizer \cite{Adam}. The robots continuously collect episodes in these environments. Upon reaching the final book page or fabric layer, the system activates a recycling mechanism to reset the environment. The policy training was completed in approximately 1700 trials across the two training environments. Hyperparameters about the policy training are in Table \ref{tab:s2}.

\begin{figure*}[!t]
    \centering
    \includegraphics[width=1\linewidth]{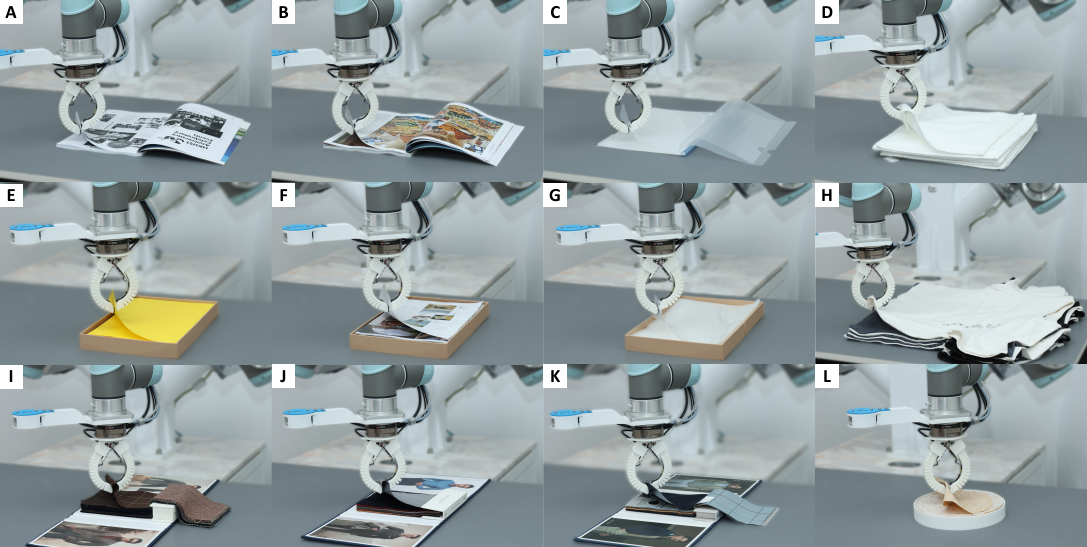}
    \caption{\textbf{Robust singulation and grasping in challenging scenarios.} The policy learned by the presented method successfully overcame a full spectrum of challenging thin, deformable objects, which include various textures of paper (\textbf{A} to \textbf{C}, \textbf{E} to \textbf{G}), suit fabrics from different seasons (\textbf{I} and \textbf{J}), hotel towels (\textbf{D}), clothing (\textbf{H}), and complex hybrid objects (\textbf{K} and \textbf{L}). Except for the book made of printer paper (A) and winter suit fabric (I), these objects had never been encountered during RL policy training.} 
    \label{fig:4}
\vspace{-0.5cm}
\end{figure*}

\begin{figure}[!t]
    \centering
    \includegraphics[width=1\linewidth]{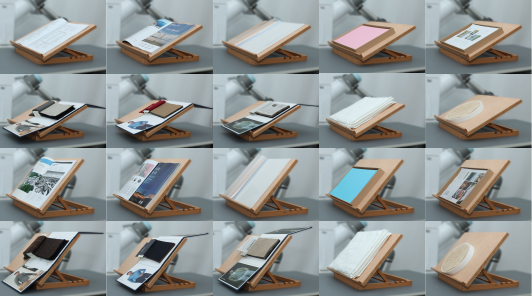}
    \caption{\textbf{Zero-shot generalization to various gravity conditions.} The inclinations of the workspace rise up to $60\degree$. As the workspace's inclinations increase, the physical interactions between layers change due to shifts in the direction of gravity. Our policy consistently demonstrated its robustness.}
    \label{fig:5}
\vspace{-18pt}
\end{figure}

\begin{figure*}
    \centering
    \includegraphics[width=1\linewidth]{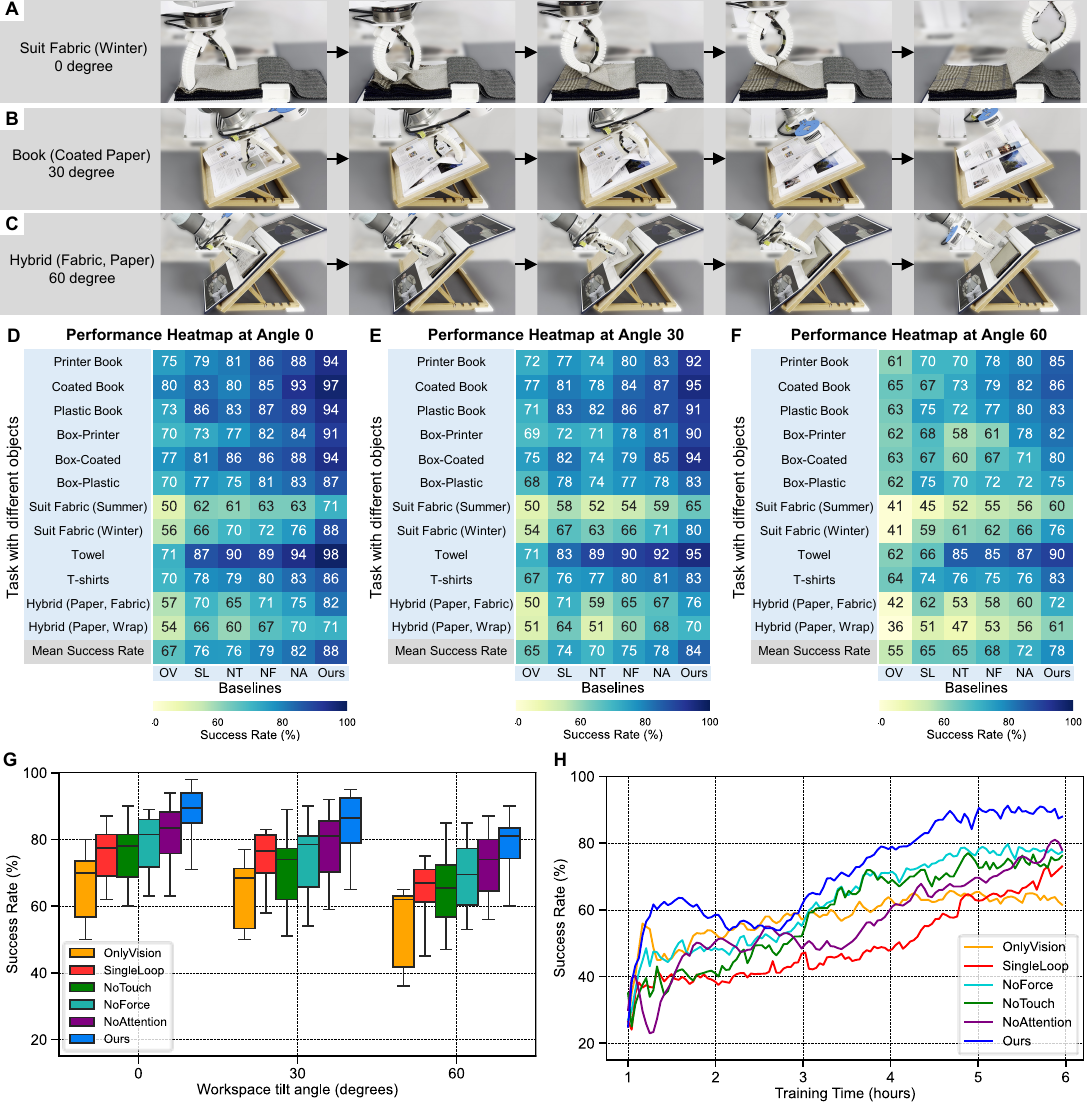}
    \caption{\textbf{Results of manipulating thin, deformable objects.} (\textbf{A} to \textbf{C}) Experiments conducted on workspaces with varying inclinations. (\textbf{D} to \textbf{E}) Performance heatmaps display task success rates for the proposed approach in comparison to ablation methods. (\textbf{F}) Effects of workspace tilt angles on manipulation performance with ablation methods, presented via a boxplot. (\textbf{G}) Training curves for manipulation policy learning across different ablation methods.}
    \label{fig:6}
\vspace{-18pt}
\end{figure*}

\section{Results}\label{sec:results}

Movie 1 summarizes the proposed approach and the experiments. We conducted policy training in two parallel environments: a book made of printer paper and a booklet crafted from winter suit fabric. These materials, with their vastly differing compressive strengths and thicknesses, cover characteristics of a broad range of thin and deformable objects. Such training objects allow robots to experience and learn from the diverse physical properties and behaviors of thin, deformable objects, ensuring the generalization of the learned policy. For all the following experiments, we utilized the same RL policy on the robot without further tuning (Fig. \ref{fig:4} and Fig. \ref{fig:5}). 

\subsection{Robust singulation and grasping across thin deformable objects}\label{sec:resultsA}

We tested our system across a variety of scenarios involving the singulation and grasping of thin, deformable objects. The thickness of these objects is shown in Table \ref{tab:s3}. These scenarios include turning book pages (Fig. \ref{fig:4}, A to C), emptying paper boxes (Fig. \ref{fig:4}, E to G), displaying fabrics (Fig. \ref{fig:4}, I and J), and picking hotel towels and clothes (Fig. \ref{fig:4}, D and H). Two hundred trials were conducted for each scenario configuration. The success of each attempt is determined by whether the robot picks only a single layer/object. The average duration per grasp is approximately 16 seconds. This includes the entire sequence: moving from the initial pose, executing the slip motion, successfully grasping and turning a single page, and returning to the starting position. The robot demonstrated consistent robustness across all scenarios (Fig. \ref{fig:6}D).

In the book page-turning task, high reflectivity and transparency of coated and plastic paper introduced noise in the depth camera, leading to imprecise robot positioning. Nevertheless, the robot's compliance enabled it to adapt to different materials and environmental characteristics, allowing successful singulation and grasping. Experiments with summer suit fabric, hotel towels, and T-shirts, each with significantly different thicknesses from the training objects, further confirmed the robot's robustness.

Hybrid materials (Fig. \ref{fig:4}, K to L) also present important challenges. The first scenario replicates a fabric booklet commonly found in suit shops, where each fabric layer is paired with a corresponding paper detailing its specifications (Fig. \ref{fig:4}K). The second scenario simulates a culinary environment where spring pancakes are interleaved by baking paper (Fig. \ref{fig:4}L). Current state-of-the-art methods require additional models \cite{flipbot} or explicit sensory data analysis \cite{l3} to manage different material types. In contrast, our approach allows the robot to perceive the unique properties of each material through the slip module and adapt policy actions accordingly directly from raw sensory data. This reduces reliance on supplementary models or extensive prior knowledge, allowing the robot to seamlessly handle novel materials.

Quantitatively, the learned policy demonstrated impressive generalization to various novel objects, achieving success rates of 97\% for coated paper, 92\% for plastic paper, 98\% for hotel towels, and 86\% for T-shirts. The success rate for mixed materials was commendable but relatively lower. This discrepancy can be attributed to two factors. First, the training process included only homogeneous materials, making generalization to mixed materials more challenging. Second, qualitative observations revealed significant variations in the friction coefficients of mixed materials. The low friction of one layer often led to the simultaneous grasping of multiple layers above it. Additionally, we observed that the layers of summer suit fabrics tended to stick together due to their material characteristics, complicating the isolation of a single layer by the robot. Further failure analysis is provided in Section \ref{sec:s3}. 

\subsection{Gravity-induced challenges}\label{sec:resultsB}

In this section, we explore how our robot singulates and grasps thin, deformable objects under the challenging conditions of different workspace tilt angles (Fig. \ref{fig:5}). Our experiments tested three tilt angles: $0\degree$, $30\degree$, and $60\degree$ (Fig. \ref{fig:6}, A to C).  We manually adjusted the robot's initial pose at different tilt angles to ensure the workspace was in view of the camera. The measured degree of workspace inclinations was only used to calculate the coordinate transformation used to perform slip motion, without affecting the decision of the manipulation policy.

As the tilt angle increases, friction between the object layers diminishes, and layers become fluffier. Our robot could handle these challenging conditions gracefully,  retaining robust manipulation (Fig. \ref{fig:6}D). The slip module enables the robot to capture changes in the object's surface and interlayer properties caused by the increased tilt angle with sensors. The learned policy adjusts actions based on multi-sensory inputs, allowing dynamic responses to changes in the environment. The robot's compliance further helps fingers conform to the object and evenly distribute contact force. Qualitative observation of the robot's behavior shows that the learned policy often tends to apply increased pressure compared to manipulation on a flat surface, thereby enhancing friction between the fingers and the object to maintain stable contact.

The success rate distribution across different tilt angles reveals a decline as the tilt angle increases (Fig. \ref{fig:6}G). This decline is attributable to the fact that higher tilt angles push the dynamics and interactions of the thin, deformable objects outside the distribution encountered during training on a flat surface. These experiments bring our robot closer to a wide range of situations encountered in real-world scenarios. See Movie 1 about how our robot helps a violinist turn sheet music placed on the stand at an angle.  

\begin{figure}[!t]
    \centering
    \includegraphics[width=1\linewidth]{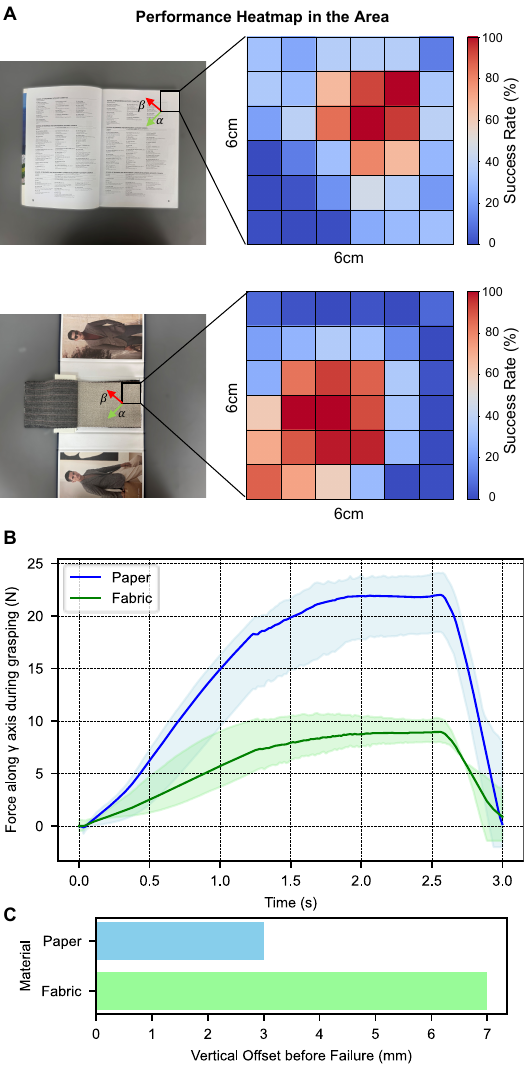}
    \caption{\textbf{Results for mechanics of the task and the robot.} \textbf{(A)} Performance heatmap showing the success rate when repeating the same grasping action but adjusting the $\alpha$-axis and $\beta$-axis offsets in the action. \textbf{(B)} Mean and range of forces along $\gamma$-axis during grasping for printer paper and winter fabric. \textbf{(C)} The range of allowable vertical uncertainty before grasp failure occurred for printer paper and winter fabric. }
    \label{fig:7}
\vspace{-18pt}
\end{figure}

\subsection{Analysis of the underlying mechanisms of the task and the robot}\label{sec:results-nx}

To investigate the mechanics of grasping thin deformable objects across different materials, we recorded a successful grasp action, and this fixed action was applied across different regions of the object. Specifically, using the upper-right corner of the object as the reference point, we incrementally adjust the $\alpha$-axis and $\beta$-axis offsets, which cover a 6x6 cm area. This region was divided into 36 individual 1x1 cm cells, and the recorded grasp action was executed 100 times at each cell. Fig. \ref{fig:7} presents the performance heatmaps for both paper and fabric materials.

For the paper material, the performance heatmap exhibits a distinct pattern with higher success rates concentrated along the diagonal (angle bisector) of the grid. This aligns with the intuitive understanding that turning a book page diagonally requires the least energy. Meanwhile, areas closer to the inner region yield very low success rates  (evident in the lower-left corner of the heatmap). This behavior reflects the inherent properties of paper, which has high compressive strength and resistance to bending. When the robot applies force away from the edge, the paper resists deformation, requiring significantly greater force to achieve sufficient bending to turn the page.

For fabric, the pattern differs. While higher success rates were also concentrated along the diagonal, the area of optimal performance shifted further from the edge. And the performance remains more consistent when moved closer to the inner regions. This behavior can be attributed to the lack of compressive strength of the fabric material, making it easier to fold or bend in the direction of the applied force.

Most importantly, this experiment reveals the critical role of our learned manipulation policy. The different distributions of grasping success rates on different materials indicate that the success cannot be determined solely by the initial pose. Instead, it depends heavily on the robot’s ability to adaptively select the appropriate grasping action based on the sensory input after the slip module provides the initial pose.  In this process, the slip module collects rich multi-sensory information. This sensory information captures detailed environmental states, such as material properties, interlayer friction. Our reinforcement learning policy then uses this multisensory input to infer and execute the optimal action for successful singulation and grasping in each specific scenario.

We also conducted experiments to evaluate the role of the soft hand in enabling robust manipulation of thin, deformable objects without requiring precise perception and control. In these experiments, a successful grasp motion was recorded and repeatedly applied to the same layer of the object, while incrementally increasing the object’s height until the grasp failed. This process was carried out on two objects: a printed book and a fabric booklet. Fig. \ref{fig:7}B depicts the mean forces along the $\gamma$-axis and their respective ranges during these grasps, as measured by the F/T sensor. Meanwhile, Fig. \ref{fig:7}C illustrates the range of allowable vertical offset before grasp failure occurred.

These results demonstrate that the soft gripper’s compliance effectively maintains the applied force within a relatively narrow range under vertical uncertainty (Fig. \ref{fig:7}B). In contrast, a rigid gripper’s response to slight vertical deviations often results in excessive forces that can trigger the arm’s safety mechanisms or damage the object. For incompressible materials like paper, this challenge is particularly pronounced. The soft gripper accommodates larger vertical offsets without compromising grasp success (Fig. \ref{fig:7}C). This capability is critical for real-world scenarios where factors like camera noise or robot accuracy cause vertical uncertainty.

\begin{table}[!t]
\centering
\vspace{0.2cm}
\caption{\textbf{Performance of position-based policies that trained on different objects without using any sensors.} Unseen objects are test cases in Fig. \ref{fig:6}D, excluding the training objects. ---$^\wedge$ stands for the object that is not used during training and counted in unseen objects.}
\begin{tblr}{
  cells = {c},
  cell{1}{1} = {r=2}{},
  cell{1}{2} = {c=3}{},
  hline{1,3,6} = {-}{},
  hline{2} = {2-4}{},
}
\textbf{Training object} & \textbf{Test object}  &                                           &                         \\
                         & Printer book & {Suit fabric\\(Winter)} & Unseen objects \\
Suit fabric (Winter)            & ---$^\wedge $                  & 47\%                                        & 43\%                    \\
Printer book     & 80\%                   &  ---$^\wedge $                                    & 46\%                    \\
Both                     & 73\%                  & 43\%                                      & 44\%                    
\end{tblr}
\vspace{-5pt}
\label{tab:nosensor}

\end{table}

\subsection{Evaluating the contribution of soft hand}\label{sec:results-compliance}

We conducted controlled experiments to quantitatively evaluate the contribution of our soft hand's passive compliance. We trained three position-based policies on different objects: printed books, suit fabric, and a combination of both. These policies mapped pre-measured surface heights directly to actions, eliminating influence from sensors. The initial hand pose was set based on the measured height, replicating the pose after the slip motion. Additionally, we also attempted to use a rigid parallel gripper to singulate and grasp thin, deformable objects with pinch grasps, but neither the position-based policies nor manually set actions were successful.

First, we tested the position-based policy trained on the suit fabric (Table \ref{tab:nosensor}, first row). During testing, each layer of fabric was often compressed due to repeated grasping, resulting in inaccurate initial gripper position and large gaps between the fingers and the object`s surface. Despite lacking sensors to perceive this, the robot achieved a 47\% success rate. In contrast, using a rigid gripper would result in losing contact with the fabric under such conditions. Our soft hand's passive compliance allows it to accommodate position inaccuracies to a certain extent and maintain continuous contact, which is critical for successful manipulation. 

We then evaluated the performance of the position-based policy trained on printed books (Table \ref{tab:nosensor}, second row). The policy demonstrated remarkable effectiveness, achieving an 80\% success rate when tested on printed books, a performance close to that of our complete system. The book pages are of uniform thickness, exhibit similar inter-layer interactions, and are difficult to compress. Using a rigid gripper to learn the policy led to collisions and emergency stops during exploration. Similarly, hard-coded actions resulted in collisions and stops due to the robot's repeatability being less than the page thickness. In contrast, the passive compliance of our soft hand absorbs minor imprecisions in operation, ensuring gentle interactions, preventing page tearing, and achieving successful singulation and grasping.

In addition, we tested the performance of a position-based policy trained using both printed books and suit fabric (Table \ref{tab:nosensor}, third row). The distinct properties of these materials led to inconsistent results when identical actions were applied at the same height. As a result, the robot struggled to learn from the reward signal reliably. Nevertheless, we achieved a certain level of success on both training and novel objects, underscoring the value of passive compliance in adapting to diverse material properties without the need for precise control.

These results show the critical role of passive compliance in successfully singulating and grasping thin, deformable objects. Although impedance control with rigid robotic hands can also be a viable method, it faces several challenging limitations. These include a strong reliance on accurate modeling of contact dynamics, increased control complexity due to the need to specify additional parameters (e.g., desired force trajectories), and significant safety concerns during real-world exploration. Our soft hand enabled the robot to perform tasks more successfully and robustly, a feat unachievable with rigid grippers. This demonstrates the principle that passive compliance can bridge the gap between precise control and the unpredictable nature of real-world interactions.

\begin{figure*}[!t]
    \centering
    \includegraphics[width=1\linewidth]{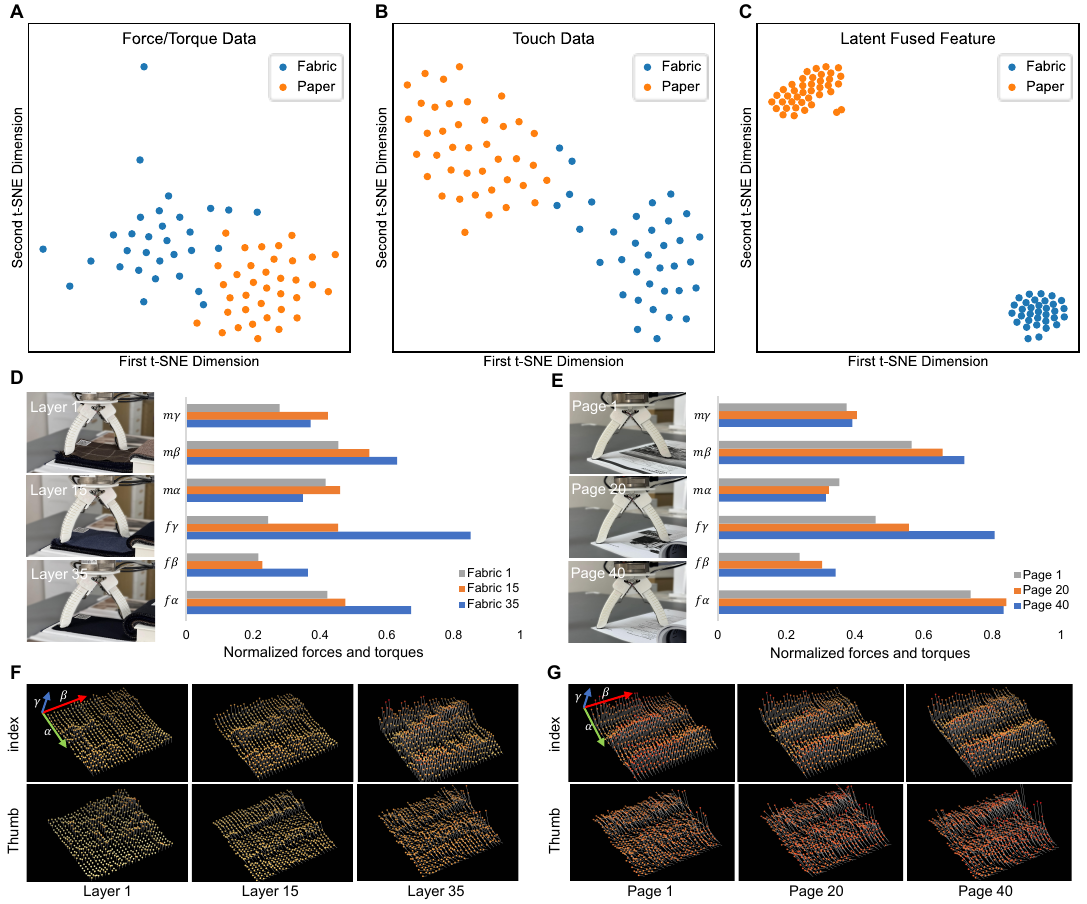}
    \caption{\textbf{Analysis and visualization of multisensory information.} (\textbf{A} to \textbf{C}) Visualization of feature clustering results by t-SNE.  (\textbf{D} and \textbf{E}) Visualization of F/T measurement after slip motion, including forces $(f{\alpha}, f{\beta}, f{\gamma})$ and torques $(m{\alpha}, m{\beta}, m{\gamma})$, on different layers of the fabric booklet and pages of the book. The force $f{\alpha}$ is parallel to the slip direction.  The force $f{\beta}$ is perpendicular to the longitudinal plane formed by the two fingers. The force $f{\gamma}$ is orthogonal to the other two forces. (\textbf{F} and \textbf{G}) Visualization of tactile responses after slip motion on different layers of the fabric booklet and pages of the book.}
    \label{fig:8}
\vspace{-18pt}
\end{figure*}

\subsection{Ablation studies}\label{sec:results-Ablation}

In the following analysis, we delve into other components of our proposed approach, including the role of tactile and F/T sensors, the effectiveness of the dual-loop learning approach, and the impact of using the attention-based encoder.

To evaluate the influence of tactile and F/T sensors, we compared our complete approach, ``Ours," with the ablation methods: ``OnlyVision" (OV), which excludes both tactile and F/T sensors during policy learning, ``NoForce" (NF), which excludes F/T sensors during policy learning, and ``NoTouch" (NT), which excludes only touch information during policy learning. 

The OV method, relying solely on depth observation, exhibits a feature bias that manifests as a heavy preference for height information. The learned policy fails to internalize the principles and mechanisms for singulating and grasping thin, deformable objects, affecting the overall performance and generalization. This limitation was reflected in the low success rate of the OV method across various objects (Fig. \ref{fig:6}D).In contrast, both NF, NT, and Ours incorporated multimodal sensors, providing a more comprehensive understanding of the environment. This allowed the learned policy to focus on critical mechanisms for task success. Quantitatively, both NT and NF underperform relative to Our method. While the tactile sensors in the NF are effective at capturing local contact events, they provide only a localized view of the contact surface. In contrast, the F/T sensor in the NT configuration provides a global proprioceptive signal that reflects distributed contact forces and torques across the entire hand-object interface, particularly those not captured directly by the fingertips. These global cues could contain critical information for inferring macroscopic object states but lack local material information. Our method integrates both modalities, thus leveraging the complementary strengths of each, as evidenced by its superior performance (Fig. \ref{fig:6}D). These findings highlight the importance of multi-modal perception in delivering comprehensive task information, thereby preventing feature dependence and facilitating effective policy learning for singulating and grasping thin, deformable objects.

Next, we evaluated our dual-loop learning structure by comparing it with a ``SingleLoop” (SL) variant, which directly explores the entire action space to learn policies. Experiments show that our method significantly outperforms the SL method within the same timeframe (Fig. \ref{fig:6}H). 

Notably, we found that the choice of action space is highly correlated with the material properties. For instance, with fabric materials used in training, a fine action space was predominantly selected. In contrast, for paper materials, the action space was more evenly distributed between coarse and fine. Table \ref{tab:s5} shows detailed statistics. This can be attributed to the differences in material surface properties and the noise characteristics of the depth camera. Fabric materials, due to their rough surfaces, generate sufficient friction with small movements from the initial pose after the slip motion. Conversely, the smooth surface of book pages necessitates larger movements to achieve the same effect. Additionally, the depth camera exhibits smaller noise on the fabric and larger noise on the paper, prompting the robot to choose a larger action space for the paper to compensate for inaccuracies in the initial positioning. The dual-loop learning structure effectively reduces the range of actions that need to be explored by first selecting the appropriate action space. This approach enhanced real-world learning efficiency by avoiding the extensive and often ineffective exploration of the entire action space as in the SL method.

We also compared the impact of using an attention mechanism to fuse sensory inputs compared to a method without it, termed ``NoAttention (NA)."  The attention mechanism, known for its ability to consider the interdependencies between different sensory inputs, creates more meaningful and context-aware feature representations. Experiment results consistently showed that our method with an attention-based multisensory encoder outperformed the NA method in all evaluation scenarios, demonstrating the effectiveness of the proposed encoder.

\subsection{Visualization the role of multisensory information}\label{sec:results-Visualization}

To examine how our system integrates different sensory streams and their respective roles, we conducted a series of controlled experiments and visualizations. We collected 40 sets of sensory inputs from different pages of a book and a fabric booklet for feature clustering. Each set included observations from tactile and F/T sensor measurements, along with fused latent features generated by the multisensory encoder.

In the t-SNE visualization for F/T sensing (Fig. \ref{fig:8}A), the close and overlapping embeddings suggest that different objects share similar properties in the F/T feature space. Fig. \ref{fig:8}(D to E) reveals a clear correlation between F/T sensing and the number of layers in the book and fabric booklet, indicating that F/T sensing primarily reflects physical interaction properties such as pressure, weight, or finger positioning, rather than the material properties.

In contrast, t-SNE visualization of touch inputs (Fig. \ref{fig:8}B) shows two relatively independent clusters, indicating that tactile sensations are more closely related to material attributes like texture and firmness. Fig. \ref{fig:8}(F to G) displays the touch data visualizations of the book and fabric booklet. The t-SNE visualization of the fused feature vector (Fig. \ref{fig:8}C) displays two independent groups with clear separation, demonstrating that integrating multisensory inputs provides a richer, more comprehensive synthesis of information than any single sensory modality.

\subsection{Discussion}\label{sec:results-Discussion}

The passive compliance of the soft hand enables robots to dexterously and reliably handle various thin, deformable materials, being robust to uncertainties while minimizing the risk of damage and reducing the need for precise perception and control. Tactile and F/T sensors equipped on the soft hand collect extensive context information about manipulation through the sliding of a finger across surfaces. The integration of multi-sensor data allows the learning process to concentrate on the core principles of the task rather than appearing feature dependence. 

To avoid complex modeling, we employ model-free RL to learn manipulation policies that drive the soft hand directly from raw sensory inputs. Learning efficiency is a critical consideration in this context. An initial pose is provided after slip motion, eliminating the need to learn how to approach the object. The proposed dual-loop learning structure reduces unnecessary exploration, and the attention-based multi-sensory encoder enhances feature fusion, both of which improve learning efficiency and performance. Additionally, rewards are autonomously determined by reading page numbers or QR code data. Automated recycling mechanisms are integrated to reset the environment, thereby minimizing the need for human intervention. 

The result is a robot of high robustness and generalization ability. The system was validated on a variety of challenging tasks that were beyond the reach of prior studies. Our controlled experiments contained 36 test scenes, blending various task scenarios, object types, and workspace tilt angles. We also demonstrated the broad potential of our method, from helping a violinist turn a music sheet to making spring pancake wraps (Movie 1). Additionally, Section \ref{sec:s2} and Movie S1 provide our further exploration regarding bag opening and cloth manipulation.

We see several limitations and opportunities for future work. First, our learned policy relies on multisensory information obtained before manipulation. This is a significant advantage, enabling one-shot manipulation and enhanced efficiency. However, the reliance on pre-obtained multisensory information has its inherent limitations. A major opportunity for future research would use the multisensory information gathered after manipulation (e.g., after the two fingers are closed). This could allow for active determination of whether multiple layers have been grasped and guide subsequent actions of the robot. Another limitation is that when one layer has significantly lower friction than the other in stacked objects, one-handed manipulation is not effective enough for singulation. Even humans will encounter difficulties in such conditions. Just as humans would use both hands to coordinate manipulation, one opportunity for future work is to learn a dual-hand system to address this situation. 

Furthermore, our current manipulation policy obtains visual perception from the depth image without fully utilizing the color image. The learned policy is independent of information such as object color, which eliminates the need to retrain or fine-tune across different environments. However, depth information ignores details present in color images that may offer insights into materials and textures. Integrating the RGB image into the network potentially improves policy performance.  Moreover, our slip module is simple and effective in predicting slip points and is learned prior to RL policy training. However, the slip module may require additional training data when faced with novel objects of different structures or shapes. This limitation stems from its inherent lack of world knowledge and inability to engage in physical reasoning. For instance, it cannot understand why a slip point would be more effective on the upper-right corner of a book than on the spine. Such knowledge usually reflects humans' high-level cognitive intelligence and is distinct from the focus of this paper, which is on low-level sensory-motor intelligence. Looking ahead, an interesting direction for future research is to leverage large foundation models \cite{llm1, llm4}  to learn high-level planning and reasoning about where to perform the slip motion. Such models have built extensive world knowledge and representations, enabling direct generalization to a wider range of objects and facilitating robot deployment in more complex environments. Additionally, our current approach demonstrates the power of passive compliance to reduce the reliance on precise perception and control. In future work, combining the passive compliance (for safety and adaptability) with impedance-like active control (for precision when needed) could further improve performance.

\section{Conclusion}

This study presents the method for learning to singulate and grasp thin, deformable objects. The learned policy drives our robot equipped with soft pneumatic grippers, achieving a significant leap over past work in terms of graspable object variety and performance. Notably, our training uses only printer paper and winter fabric on flat surfaces. However, the robot successfully handles various deformable objects with different physical properties (coated paper, hotel towels, T-shirts), hybrid materials (fabrics separated by pages or spring pancakes separated by baking paper), and gravitational effects. Our results suggest that the extraordinary complexity of manipulating thin, deformable objects can be tamed with embodied multisensory integration and compliance in the soft gripper. By enabling the robot to adapt to the complex dynamics of deformable objects without relying on precise control, our approach opens new possibilities for robots to operate in unstructured environments and perform tasks that require delicate handling. Future work may involve extending this framework to more complex manipulation tasks and further improving the efficiency and generalization of the learning process.




\section{Appendix}
\textbf{This Appendix includes:}

Sections S1 to S10

Figs. S1 to S4

Tables S1 to S7

Other Supplementary materials, including Movie 1, Movies S1 to S2, part of the hardware models, instructions, data, and code, will be at \url{https://robotll.github.io/LTDOM/}






\setcounter{figure}{0}
\renewcommand{\figurename}{Fig.}
\renewcommand{\thefigure}{S\arabic{figure}}
\renewcommand{\theHfigure}{Supplement.\thefigure}

\captionsetup[table]{labelfont={bf},labelformat={default},name={Table}}
\setcounter{table}{0}
\renewcommand{\thetable}{S\arabic{table}}
\renewcommand{\theHtable}{Supplement.\thetable}

\setcounter{section}{0}
\renewcommand{\thesection}{S\arabic{section}}
\renewcommand{\theHsection}{Supplement.\thesection}

\begin{figure*}[!ht]
    \centering
    \includegraphics[width=1\linewidth]{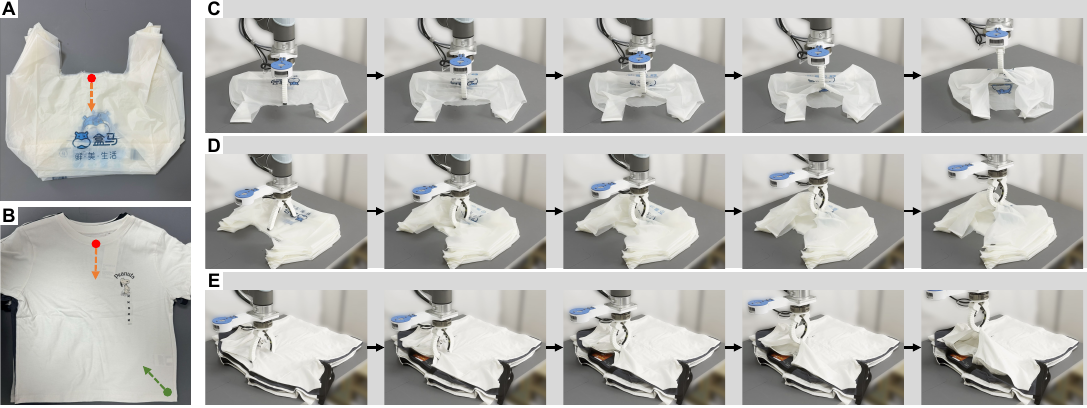}
    \caption{\textbf{Additional experiments on bag opening and garment handling.} (\textbf{A}) Annotation of slip point and its direction on the bag. (\textbf{B}) The red line indicates the new annotation of the slip point and its direction, while the old annotation is shown as the green line. (\textbf{C} and \textbf{D}) Illustration of the bag opening task, where a single or multiple bags are placed on a table. (\textbf{E}) Illustration of singulating and grasping garments, incorporating updated human preferences.}
    \label{fig:s1}
\vspace{-5pt}
\end{figure*}

\section{Nomenclature}\label{sec:s1}

$S$ \qquad \ \ state space 

$A$ \qquad \ \  action space 

$R$ \qquad \ \ reward function 

$P$ \qquad \ \ transition probability 

$a$ \qquad \ \ action 

$r$ \qquad \ \ reward 

$s$ \qquad \ \ state

$o$ \qquad \ \ observation

$\lambda$ \qquad \ \ discount factor

$\pi$ \qquad \ \ policy

$(\cdot)^{out}$ \quad outer loop quantity

$(\cdot)^{in}$ \quad \ inner loop quantity

$(\cdot)_{t}$ \quad \ \ \ quantity at time step $t$

$f(\cdot)$ \quad \ \ force quantity

$m(\cdot)$ \quad \ torque quantity

$g(\cdot)$ \quad \ \ encoder

$l(\cdot)$ \quad \ \ \ latent vector

$(\cdot)^{vis}$ \quad visual quantity

$(\cdot)^{ind}$ \quad index touch quantity

$(\cdot)^{thu}$ \quad thumb touch quantity
 
$(\cdot)^{pro}$ \quad muscle proprioception quantity

$(\cdot)^{aux}$ \ \ \  auxiliary state quantity

$(\cdot)^{mul}$ \ \ \  multisensory quantity

$(\cdot)^{c}$ \quad \ \ \ cross-attention quantity

$(\cdot)^{\kappa}$ \quad \ \ \ transformer quantity

$H(\cdot)$ \quad \  coordinate system

${E}(\cdot)$ \quad \ expectation 

$\Lambda$ \qquad \ \ longitudinal plane formed by fingers

$\alpha$ \qquad \ \ axis parallel to slip direction and $\alpha \in \Lambda$

$\gamma$ \qquad \ \ axis $\gamma \perp \alpha$ and $\gamma \in \Lambda$; 

$\beta$ \qquad \ \ axis $\beta \perp (\alpha \times \gamma)$; 

$x$ \qquad \ \ displacement along $\alpha$  

$z$ \qquad \ \ displacement along $\gamma$

$\theta$ \qquad \ \ orientation about $\beta$

\section{Additional experiments on garments and bag opening}\label{sec:s2}

To further validate the robustness of our learned policy and the flexibility of our method, we conducted two supplemental experiments.  Both experiments employed the previously learned RL policy without fine-tuning, merely adding extra data to retrain the slip module.

Our first experiment investigated the task of bag opening.  To adapt the slip module to this task, we additionally annotated data regarding slip points and directions on plastic bags (Fig. \ref{fig:s1}A). In the experimental setup, we considered two distinct scenarios: one involving a single plastic bag (Fig. \ref{fig:s1}C) and another with a stack of them (Fig. \ref{fig:s1}D). In both scenarios, the robot was required to singulate a single bag layer. 

Furthermore, different tasks may prompt robots to have different preferences for grasping clothing, such as opting to hang clothes by grasping the collar.  Here, we replaced the old slip annotation (Fig. \ref{fig:s1}B green line)  on the cloth with a new one (Fig. \ref{fig:s1}B red line), making the slip module learn a different preference. This adjustment demonstrates the flexibility of the slip module (Fig. \ref{fig:s1}E).

Quantitative results from both sets of experiments are presented in Table \ref{tab:s1}. Notably, the above test settings significantly extend beyond the scope of existing literature on deformable object manipulation, which usually assumes that only a piece of clothing or a bag is on the table. The outcomes of these exploratory trials underscore the robustness and adaptability of our methodology, demonstrating its promise for real-world applications in clothing and bag manipulations, even under complex conditions.

\begin{table}[!t]
\centering
\caption{Additional experiments on cloth and bag opening}
\begin{tblr}{
  cells = {c},
  hline{1,3} = {-}{0.08em},
}
                  & Single Bag & Multi-bags & Cloth \\
Successful Trials & 99/100      & 63/100      & 81/100 
\end{tblr}
\label{tab:s1}
\end{table}

\begin{figure*}[!t]
    \centering
    \includegraphics[width=1\linewidth]{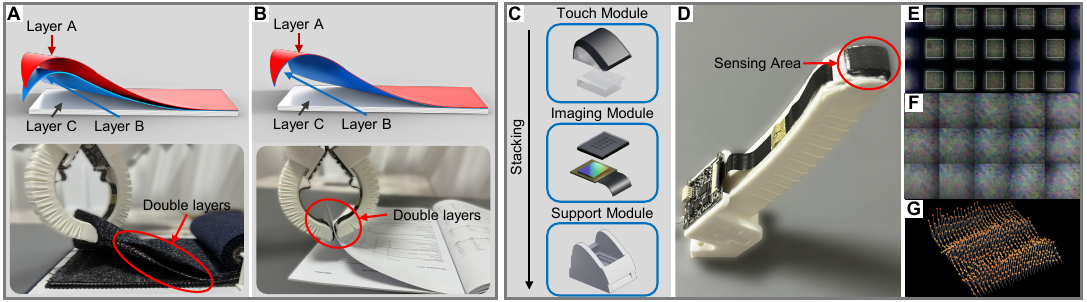}
    \caption{\textbf{Analysis of failure cases and touch sensor design.} (\textbf{A}) A failure case where the friction between layer B and layer C is so low that both layer A and layer B lift. (\textbf{B}) A failure case where layer A adheres to layer B, resulting in the unintended lifting of both layers. (\textbf{C}) The design of our touch sensor, which includes the touch, imaging, and support modules. (\textbf{D}) This image shows the robotic gripper equipped with the touch sensor, highlighting the sensing area. (\textbf{E}) Raw image from the sensor. (\textbf{F}) Stitched image created from 15 tiles of the raw image. (\textbf{G}) The final deformation field.}
    \label{fig:s2}
\vspace{-5pt}
\end{figure*}

\section{Failure analysis}\label{sec:s3}

Our experiments encountered several cases where the robot failed to achieve the desired singulation and grasping. The most frequent failure was the unintentional grasping of multiple layers. We observed two primary causes for this form of failure. 

The first cause relates to a low friction coefficient between certain object layers. As demonstrated in Fig. \ref{fig:s2}A, when Layer B or C has an exceptionally smooth surface, the frictional forces between them are significantly less than those between layers A and B. As a result, layers A and B are simultaneously grasped. This phenomenon is primarily observed in scenarios involving fabrics and mixed materials, which is due to the variations in surface smoothness among fabrics and the natural disparity in friction coefficients among different materials. 

The second cause of failure, as illustrated in Fig. \ref{fig:s2}B, arises when layers are stuck together. This is commonly observed in contexts involving paper, where even manual separation requires considerable effort. The adhesion often results from microscopic structural factors or external variables like ink, printing, and Van der Waals forces. Under such conditions, the frictional force between layers A and B is significantly greater than between layers B and C, leading to multiple layers being grasped together. As discussed in the experiments of the book page turning with printer paper, we attempted 200 flips and succeeded 188 times, yielding a success rate of 94\%. Among the 12 unsuccessful attempts, 11 were attributed to tight adhesion between pages and one to noise errors in the depth camera reading.

In addition, we observed objects slipping out of the fingers during manipulation due to camera noise or inappropriate action prediction, especially in scenes not encountered during training. Moreover, when the workspace has a high tilt angle, it is easier to grasp multiple layers due to the altered friction dynamics.  Movie S2 shows the representative failure cases during experiments.

\section{Touch sensor architecture}\label{sec:s4}

We used a vision-based tactile sensor of fingertip dimensions based on \cite{OurTouch}. While sensors like GelSight \cite{Gelsight} and DIGIT \cite{DIGIT} can offer high-resolution tactile feedback, they are constrained by their larger dimensions and flat geometry; hence, they are hard to integrate into a space the size of a human fingertip. In contrast, our sensor is able to provide a large field of view in a compact form factor, benefiting from the compound-eye imaging system \cite{OurTouch}, making it ideal for integration into the minuscule fingertips of soft grippers.

Our tactile sensor consists of three crucial components, as illustrated in Fig. \ref{fig:s2}C: the touch, imaging, and support modules. The touch module features a curved elastomer layer designed to conform to the fingertip of our soft gripper, and an acrylic plate. The elastomer layer is constructed by a transparent layer embedded with a random color pattern, a white protective layer, and a black layer designed to stabilize light intensity variations. The imaging sub-system beneath the curved touching layers includes a microlens array, a pinhole structure to define the aperture, and a Complementary Metal Oxide Semiconductor (CMOS) image sensor. Enclosing these is the support module, a 3D-printed shell that provides the structural integrity required for the effective functioning of the tactile sensor. Fig. \ref{fig:s2}D displays the sensor when mounted on the soft finger. We obtain the touch information with the following steps. First, we crop 15 tiles from the raw image (Fig. \ref{fig:s2}E), which are subsequently stitched together (Fig. \ref{fig:s2}F). Next, we employ the same algorithm in \cite{OurTouch} to process the combined image and obtain the deformation field (Fig. \ref{fig:s2}G).

\section{Parameters of observation and action}\label{sec:s5}

The observation vectors are defined in Table \ref{tab:s4}. Visual observation $o_t^{vis}$ is defined as a $40\times40$ depth image cropped from the region surrounding the slip point. Tactile information $o_t^{ind}, o_t^{thu}$ at each fingertip is encapsulated in a $25\times25\times3$ matrix, which represents the deformation field in the $\alpha$, $\beta$, and $\gamma$ directions, and samples 625 points across the tactile area. 

In the fine-grained action space of the inner loop, \( z_t \) ranges from -3 mm to 3 mm at 1 mm intervals, approximating the thickness of 60 sheets of printer paper or 10 layers of winter fabric. \( x_t \) varies between -7.5 mm and 7.5 mm, with a 5 mm interval, and \( \theta_t \) is confined between 0\degree{} and 3\degree{}, at 1\degree{} intervals. Conversely, the ranges in the coarse action space are 2.5 times larger than those in the fine action space, with the same discrete number. Therefore, \( z_t \) extends from -7.5 mm to 7.5 mm, with a 1 mm interval, \( x_t \) spans from -18.7 mm to 18.7 mm, with a 5 mm interval, and \( \theta_t \) ranges between 0\degree{} and 7.5\degree{}, at 1\degree{} interval. 

\begin{figure*}[t]
    \centering
    \includegraphics[width=1\linewidth]{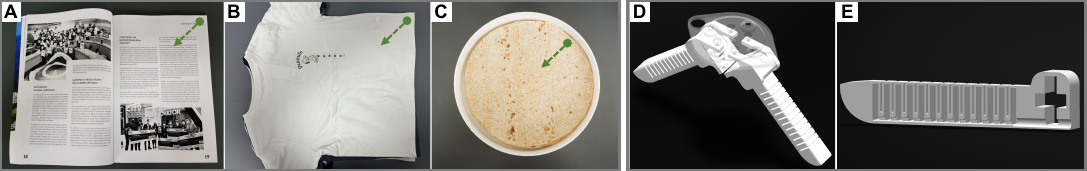}
    \caption{\textbf{Annotated images and models of robot hand.} (\textbf{A to C}) Annotation on three kinds of objects used for learning the slip network. (\textbf{D}) 3D model for the two-fingered soft robot hand. (\textbf{E}) Internal air chamber structure of the finger.}
    \label{fig:s3}
\vspace{-15pt}
\end{figure*}

\section{Slip network architecture}\label{sec:s6}

The network within the slip module is designed as a fully convolutional neural network (FCN). Specifically, we discretize the image rotation within a range of $(-90^{\circ}, 90^{\circ})$ at intervals of $15^{\circ}$, corresponding to slip directions in the environment. To ensure the object remains within the frame post-rotation, we pad the image prior to adjusting its resolution to $640 \times 480 \times 1$. Then the slip network utilizes a ResNet-18 backbone \cite{Resnet} for feature extraction. Upon extracting the features, the network incorporates three $1 \times 1$ convolution layers. The initial layer reduces the 512 channels from ResNet-18 outputs to 128 channels, followed by 2x bilinear upsampling. The next layer further reduces the channels to 32, followed by another 2x bilinear upsampling. The last layer maps these 32 channels to a $640 \times 480$ probability map. The slip network is optimized using the Adam optimizer \cite{Adam}, with a learning rate of 0.001 and a binary cross-entropy loss function. The training dataset is extended by data augmentation from manually annotated images (Fig. \ref{fig:s3} A to C). 

\begin{table}[t]
\centering
\caption{Dimensionality of observations.}
\begin{tblr}{
  cell{1}{1} = {c=2}{},
  cell{3}{1} = {r=2}{},
  vline{2} = {1}{},
  vline{3} = {2-6}{},
  hline{1-2,7} = {-}{},
}
Input                 &              & Dimensionality \\
visual observation    & $o_t^{vis}$    & 40x40          \\
tactile observation   & $o_t^{ind}$    & 25x25x3        \\
                      & $o_t^{thu}$    & 25x25x3        \\
muscle proprioception & $o_t^{pro}$    & 6              \\
auxiliary information & $o_t^{aux}$ & 2              
\end{tblr}
\label{tab:s4}
\end{table}

\begin{table}[t]
\centering
\caption{Hyperparameters for SAC}
\begin{tabular}{rr} 
\toprule
\multicolumn{1}{l}{number of environments} & 2                      \\
learning rate                              & 0.003                  \\
gradient steps                             & 3                      \\
buffer size                                & 10000                  \\
learning starts                            & 10                     \\
gamma                                      & 0.99                   \\
batch size                                 & 64                     \\
\bottomrule
\end{tabular}
\label{tab:s2}
\vspace{-5pt}
\end{table}

\section{Network architecture of policy}\label{sec:s7}

For base encoders, the visual encoder $g_{\text{vis}}$  is designed as a Convolutional Neural Network (CNN), consisting of two convolutional layers with 16 and 32 filters, each employing a 7x7 kernel and a stride of 2. The layers are followed by Rectified Linear Unit (ReLU) \cite{RELU} activation functions and an adaptive average pooling layer. The touch encoder $g_{\text{ind}}$ and $g_{\text{thu}}$ are also modeled as CNNs, mirroring the visual encoder in kernel size, stride, and padding. The muscle proprioception encoder $g_{\text{pro}}$ and auxiliary state encoder $g_{\text{aux}}$ are constructed as linear layers containing 32 hidden units. In the initial phase, each observation element is transformed into a 32-dimensional latent vector using base encoders, which are then fed into the multisensory encoder. 

The multi-sensor encoder consists of a cross-attention block $g_{c}$ and a transformer encoder $g_{\kappa}$. $g_{c}$ is composed of two identical cross-attention layers \cite{CrossAttention}. $g_{\kappa}$ is composed of two transformer layers  \cite{Attention}, each featuring eight attention heads and a feedforward network with 64 hidden dimensions. We removed the position embedding, as in our case, the fusion of features is independent of the input order of the features. 

Finally, the 64-dimensional output of the transformer encoder is fed into the MLP. The final MLP block has two hidden layers with $\{64, 1\}$ hidden units for the outer loop and $\{64, 4\}$ hidden units for the inner loop. The details of the hyperparameters for policy training can be found in Table \ref{tab:s2}.

\begin{figure}[!ht]
    \centering
    \includegraphics[width=1\linewidth]{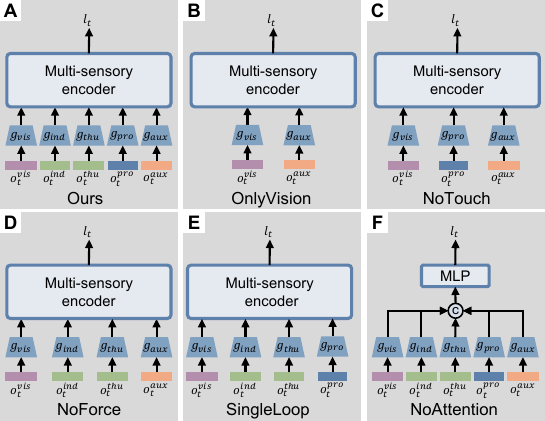}
    \caption{\textbf{Comparison of ablation model architectures.} (\textbf{A}) Our method architecture incorporates inputs from various sensor modalities. (\textbf{B}) OnlyVision architecture relies solely on visual input. (\textbf{C}) NoTouch architecture excludes touch input but includes other modalities. (\textbf{D}) NoForce architecture excludes the F/T sensor input but includes other modalities. (\textbf{E}) SingleLoop architecture is a version without the inner-outer loop learning process; thus, it does not incorporate auxiliary status inputs. (\textbf{F}) NoAttention architecture employs an MLP without the attention mechanism for processing sensor inputs.}
    \label{fig:s4}
\vspace{-15pt}
\end{figure}

\section{Ablation model architectures}\label{sec:s8}

This section provides additional details on the four ablation methods used in our evaluations. Comparisons involving these ablation models and our method are presented in Fig. \ref{fig:s4}.

\begin{itemize}
    \item \textbf{OnlyVision (OV):} This approach only relies on visual observation to learn the manipulation policy, as shown in Fig. \ref{fig:s4}B. The objective is to evaluate whether visual information is sufficient for completing thin deformable object manipulation;
    
    \item \textbf{NoTouch (NT):} The observation is exclusively based on visual and muscle proprioceptive perception to validate the importance of touch information. Refer to Fig. \ref{fig:s4}C for a visual representation of this ablation model architecture;

    \item \textbf{NoForce (NF):} The observation is exclusively based on visual and tactile perception to validate the importance of force and torque information. Refer to Fig. \ref{fig:s4}D for a visual representation of this ablation model architecture;
    
    \item \textbf{SingleLoop (SL):} This method only uses the inner loop to learn manipulation policy in the entire action space, eliminating the need for auxiliary observation input, as depicted in Fig. \ref{fig:s4}E. Specifically, the range for \(z_t\) is from -7.5 mm to 7.5 mm, with an interval of 1 mm. \(x_t\) varies between -18.7 mm and 18.7 mm, with a step size of 5 mm, while \( \theta_t \) is confined between $0\degree$ and $7.5\degree$, spaced at $1\degree$ intervals. This ablation model serves to assess the efficiency of the proposed dual-loop learning structure under real-world training conditions;  
    
    \item \textbf{NoAttention (NA):} We concatenate the latent feature vectors from the basic encoder and replace the multisensory encoder with an MLP block (Fig. \ref{fig:s4}F). This setup aims to evaluate the effectiveness of the proposed attention-based multisensory encoder when processing multisensory information.
\end{itemize}

\begin{table*}[!t]
\centering
\caption{Comparison of our gripper with prior designs for thin-object manipulation and general soft robotic hands.}
\resizebox{\textwidth}{!}{%
\begin{tblr}{
  cells = {c},
  vline{2} = {2-7}{},
  hline{1-2,8} = {-}{},
}
Feature                & Ours                     & Flipbot\cite{flipbot} & RoTipBot\cite{revision_R2_1}             & Tirumala et al.\cite{l3}                   & Zhang et al.\cite{soft1}       & Truby et al.\cite{soft2}  & Puhlmann et al.\cite{soft3}        \\
Gripper type       & Soft                   & Soft    & Rigid w/ soft tactile             & Rigid             & Soft                         & Soft         & Soft             \\
Actuation          & Pneumatic                   & Pneumatic    & Motorized         & Motorized         & Pneumatic                         & Pneumatic         & Pneumatic             \\
Compliance         & High                   & High    & Low               & Low               & High                         & High         & High             \\
Control complexity & Low                    & Low     & Medium            & Low               & High                         & Medium         & High           \\
Integrated sensors & Fingertip tactile, F/T & F/T     & Fingertip tactile & Fingertip tactile & Palm tactile~ ~ & Ionogel      & - \\
Soft structure fabrication      & Easy                   & Easy    & -                 & -                 & Complex                      & Complex      & Medium           
\end{tblr}
}
\label{tab:gripper}
\vspace{-5pt}
\end{table*}

\section{Novelty and Rationale of the Gripper Design}\label{sec:s9}

Our gripper design is based on the well-established fast-PneuNet \cite{revision_R1_4} architecture but uniquely integrates several key features tailored explicitly for manipulating thin, deformable objects. These key features include underactuation, passive compliance, and multimodal sensing via fingertip tactile sensors combined with a wrist-mounted force/torque sensor. 

Compared to other grippers used in literature for grasping thin, deformable objects, while \cite{revision_R2_1, l3} utilizes fingertip tactile sensors as ours, its rigid gripper lacks the whole-finger passive compliance intrinsic to our soft fingers. We have empirically demonstrated in our experiments (Section IV-D) the substantial contribution of passive compliance in the successful manipulation of thin, deformable objects.

Furthermore, in comparison to other soft robotic hands, our design emphasizes simplicity and practical usability. Each finger utilizes only a single pneumatic channel for bending. This underactuation significantly reduces the control complexity,  making it far tractable than controlling multiple joints in a rigid or more complex multi-chambered soft hand.

For example, \cite{soft1} focuses on achieving human-like dexterity through coordinated palm-finger movements and high-density tactile sensing in the palm. Although highly capable, such control complexity is not strictly necessary for the singulation and grasping of thin, deformable objects. Similarly, \cite{soft2} relies on discrete, complex actuation modes that significantly increase fabrication and control overhead. Given our model-free RL framework, which learns directly in the real world, minimizing control complexity is crucial for efficient training and generalization. Our hand strikes a balance between sensory richness and control simplicity, enabling effective performance without over-engineering.

Another critical advantage of our gripper is its manufacturability and accessibility. The soft fingers can be fabricated within approximately 12 hours using consumer-grade 3D printers (e.g., Bambu P1P, priced approximately \$500-\$700) and widely available thermoplastic polyurethane (TPU) filament (approximately \$40). This straightforward manufacturing approach eliminates the need for specialized processes like silicone casting \cite{soft1} or embedded 3D printing \cite{soft2}, significantly reducing barriers for adoption within the robotics research community. While we chose to integrate our own fingertip tactile sensors for convenience, it is worth noting that the geometry of the finger is also easily modifiable due to its reproducibility and ease of fabrication, allowing for easy adaptation to a variety of commercially available or custom sensors as needed.

In summary, the novelty of our gripper design lies in its specific combination of underactuation, passive compliance, and multimodal sensory integration, all tailored for the challenges of manipulating thin deformable objects using reinforcement learning. This unique integration reduces control complexity, enhances robustness and adaptability, provides rich sensory feedback essential for our slip module, and supports the "imprecise dexterity" required for manipulating thin, deformable objects. Detailed comparisons across different robotic hands are summarized in Table \ref{tab:gripper}.

\begin{table}[!t]
\centering
\caption{Success rates (\%) across different object sizes. The medium size corresponds to the objects utilized in the experiments presented in the main text.}
\vspace{0.2cm}
\resizebox{0.48\textwidth}{!}{%
\begin{tblr}{
  column{even} = {c},
  column{3} = {c},
  hline{1-2,5} = {-}{},
}
Object Type                    & Small & Medium & Large \\
Printer Paper (A6, A4, A2)     & 90  & 91   & 89  \\
Towel (15x15, 30x30, 45x45 cm) & 98  & 98   & 97  \\
T-shirt (XXS, S, L)            & 84  & 86   & 87  
\end{tblr}
}
\label{tab:objectsize}
\vspace{-5pt}
\end{table}

\begin{table}[!t]
\centering
\caption{Object thickness in experiments (mm)}
\begin{tabular}{ll} 
\toprule
printer paper      & 0.05-0.1  \\
coated paper       & 0.05      \\
plastic paper      & 0.15      \\
winter suit fabric & 0.3-0.7   \\
summer suit fabric & 0.2-0.4  \\
hotel towel              & 1.5       \\
cloth              & 0.7       \\
baking paper       & 0.025     \\
pancake wrap       & 1         \\
\bottomrule
\end{tabular}
\label{tab:s3}
\vspace{-5pt}
\end{table}

\begin{table}[!t]
\centering
\caption{Proportion of action space selection}
\begin{tblr}{
  width = \linewidth,
  colspec = {Q[294]Q[250]Q[371]},
  cells = {c},
  cell{1}{1} = {r=2}{},
  cell{1}{2} = {c=2}{0.621\linewidth},
  hline{1,5} = {-}{0.08em},
  hline{3} = {2-3}{},
}
Selected space & Objects      &                      \\
               & Printer book & Suit fabric (Winter) \\
Coarse~        & 44\%           & 13\%                   \\
Fine           & 56\%           & 87\%                   
\end{tblr}
\label{tab:s5}
\vspace{-5pt}
\end{table}

\section{Discussion of action space granularity generalization}\label{sec:s10}

A pertinent question regarding the dual-loop learning framework presented in this work is whether its chosen coarse and fine action space granularities are effective only for the particular type and size of thin deformable objects tested in the experiments, or if they have broader applicability. 

The defined action space granularities can generalize across various object sizes due to the inherently localized nature of interactions involved in manipulating thin, deformable objects. Analogous to human manipulation tasks, such as page-turning, interactions primarily occur at specific localized regions (e.g., corners or edges) regardless of the overall object dimensions. Hence, substantial variations in object sizes minimally affect the fundamental localized deformation and interaction dynamics, allowing the established coarse and fine action granularities to remain effective.

To quantitatively validate this assertion, we performed supplementary experiments involving objects of diverse dimensions, including printer paper sheets (A2, A4, A6), hotel towels (15 cm × 15 cm, 30 cm × 30 cm, 45 cm × 45 cm), and T-shirts (XXS, S, L). Results summarized in Table \ref{tab:objectsize} demonstrate that size variations minimally influence the effectiveness of the chosen action granularities. For example, the page-turning task consistently achieved high success rates (approximately 90\%) across all paper sizes. Comparable success rates were observed in grasping tasks involving towels and T-shirts, confirming robust generalization across a broad spectrum of object sizes.

In principle, the current action granularity should remain effective for any unseen object whose physical properties fall within the range of the materials we tested (e.g., similar surface friction or bending stiffness). Our policy does not rely on predefined strategies for specific materials but instead learns to associate different sensory signals with the required actions (or action granularities).

The learned policy used for testing is trained solely on printer paper and winter suit fabric, all without any modification to granularity.  As shown in our results, the granularity settings have exhibited strong generalization across various unseen object types, including coated paper, plastic paper, summer suit fabrics, hotel towels, T-shirts, and hybrid paper/fabric stacks. These objects span a significant range of thicknesses, stiffnesses, surface textures, and friction coefficients, representing typical thin deformable objects encountered daily. This empirical evidence supports the hypothesis that, for unseen objects with similar physical properties to those we tested, our action space granularities will remain effective. 

\bibliographystyle{IEEEtran}
\bibliography{references}

\end{document}